\documentclass[Afour,sageh,times]{sagej_arxiv}
\usepackage{moreverb,url}
\usepackage{multirow}
\usepackage{algorithmic}
\usepackage{algorithm}
\usepackage[colorlinks,bookmarksopen,bookmarksnumbered,citecolor=red,urlcolor=red]{hyperref}
\usepackage{orcidlink}
\newcommand\BibTeX{{\rmfamily B\kern-.05em \textsc{i\kern-.025em b}\kern-.08em
		T\kern-.1667em\lower.7ex\hbox{E}\kern-.125emX}}

\begin{document}
	
	\runninghead{Sabzejou and Azimmohseni}
	
	\title{Skeleton-Guided Geometry-Aware Auto-Labeling for Robotic Grasping: From 4-DoF Learning to 6-DoF Execution}
	
	\author{Ali Sabzejou\affilnum{$^*$}\affilnum{1}, Farbod Azimmohseni\affilnum{$^*$}\affilnum{1}, Mehdi Tale Masouleh\affilnum{1}, Ahmad Kalhor\affilnum{1}}
	
	\affiliation{\affilnum{1}Human and Robot Interaction Laboratory, School of Electrical and Computer Engineering, University of Tehran, Tehran, Iran\\
		\affilnum{$^*$} Equal Contribution}
	
	\corrauth{Mehdi Tale Masouleh,
		Human and Robot Interaction Laboratory,
		School of Electrical and Computer Engineering,
		University of Tehran,
		Tehran,
		Iran}
	
	\email{m.t.masouleh@ut.ac.ir}

% Journal-version notice in the first-page author footnotes.
% The Sage-style title footnotes are placed in the right column.
\makeatletter
\g@addto@macro\@email{%
\footnotetext[0]{\sagesf
This is the authors' accepted manuscript of the article
``Skeleton-Guided Geometry-Aware Auto-Labeling for Robotic Grasping:
From 4-DoF Learning to 6-DoF Execution,''
published in \textit{The International Journal of Robotics Research}
(IJRR), 2026. The final published version is available at
\url{https://doi.org/10.1177/02783649261477772}.}%
}
\makeatother
	
	\begin{abstract}
		Robotic grasping relies on large manually annotated datasets, which are costly to create. This work introduces a novel algorithm for generating two-finger grasp keypoints based on straight skeletons, identifying high-potential regions for stable and feasible grasp pose estimation. An automated pipeline is developed to perform grasp auto-labeling without human intervention, facilitating large-scale annotation with consistent quality. Furthermore, an architecture for 4-DoF grasping, named the Skeleton-based Generative Grasp Convolutional Neural Network (SkelGG-CNN), is introduced, which incorporates skeleton-based guidance during training to directly learn geometric grasping features. 
Finally, the developed auto-labeling is extended to 6-DoF grasping and applied directly to point clouds via a generalized quasi-3D skeletonization algorithm. 
Evaluations demonstrate that SkelGG-CNN achieves accuracies of 95.42\%, 96.55\%, and 97.1\% on the Jacquard V1, Jacquard V2, and Cornell benchmarks, respectively, using 4-DoF auto-generated labels. On GraspNet, the 6-DoF auto-labeling algorithm yields an average precision of 18.33 on the novel split, comparable to state-of-the-art model-based approaches. In simulation, the generalized 6-DoF auto-labeling method yields success rates of 98.4\% on the Dex-Net test set and 89.1\% on EGAD!. Real-world experiments on a Delta parallel robot confirm 97.4\% success on single household objects and 95.0\% in cluttered scenes, outperforming state-of-the-art methods without domain adaptation. Experiments show that the framework's 4-DoF pipeline is adaptable across environments, generating grasp annotations that approximate human-level precision, enabling efficient training of lightweight models like SkelGG-CNN on small datasets for real-world deployment. Additionally, the generalized 6-DoF auto-labeling algorithm extends this automation to point clouds, providing scalable, geometry-aware annotations for cluttered 3D scenes.
		Source code and videos are available at \url{https://github.com/Farbod82/SkelGGCNN}.
	\end{abstract}
	
	
	\maketitle

	\section{Introduction}
	Robotic manipulation has been a significant area of research, with the task of grasping objects emerging as one of the most widely studied topics in recent decades.
	Although grasping appears simple to humans, it remains a difficult task for robots, and various approaches have been explored to address it.
	The difficulty lies in determining appropriate grasp poses for a diverse range of objects, each differing in geometry, gripping force requirements, and weight distribution.
	These factors significantly influence how robotic systems effectively grasp objects from specific configurations, making the task highly complex and context-dependent.
	In early studies, analytical methods were the primary tools used to determine grasp poses, focusing on geometric approaches to analyze the structure and properties of objects.
	These methods typically relied on mathematical models and physical principles, such as force closure and friction modeling, to compute stable and reliable grasp configurations \cite{b19}.
	While effective for simple and well-defined objects, these techniques often struggle in complex or unknown scenarios.
	Additionally, these methods tend to be computationally intensive and lack the ability to generalize effectively in dynamic environments.
	
	The emergence of deep neural networks and data-driven methods in computer vision has significantly advanced the ability to tackle the complex challenges of robotic grasping \cite{b20}.
	By leveraging image processing techniques originally developed for computer vision, grasping models have become increasingly robust, achieving strong performance in unfamiliar environments and real-world applications.
	With the help of large-scale grasping datasets, these models can generalize effectively to handle objects with diverse shapes, physical properties, and environmental conditions, resulting in enhanced reliability in grasping tasks.
	However, despite the impressive capabilities of these models, significant challenges persist. 
	One of the most critical issues is the heavy reliance of these data-driven methods on annotated datasets for training.
	Creating these datasets often requires substantial human effort, involving extensive manual labeling of grasping points.
	This dependence on extensive labeled data not only increases development costs but also limits the adaptability of these models to new, unforeseen tasks or environments where labeled data may be scarce or unavailable \cite{wang2021demograsp, b63}.
	Hybrid methods that combine analytical grasp pose estimation with learning-based refinement have been explored extensively in prior research.
	These methods mostly rely on random sampling methods for initial grasp generation, which requires extensive filtering to acquire high-quality grasp data~\cite{Pointnetgpd, mousavian20196, b60}.
	
\begin{figure}[!t]
		\centering
		\includegraphics[width=\linewidth]{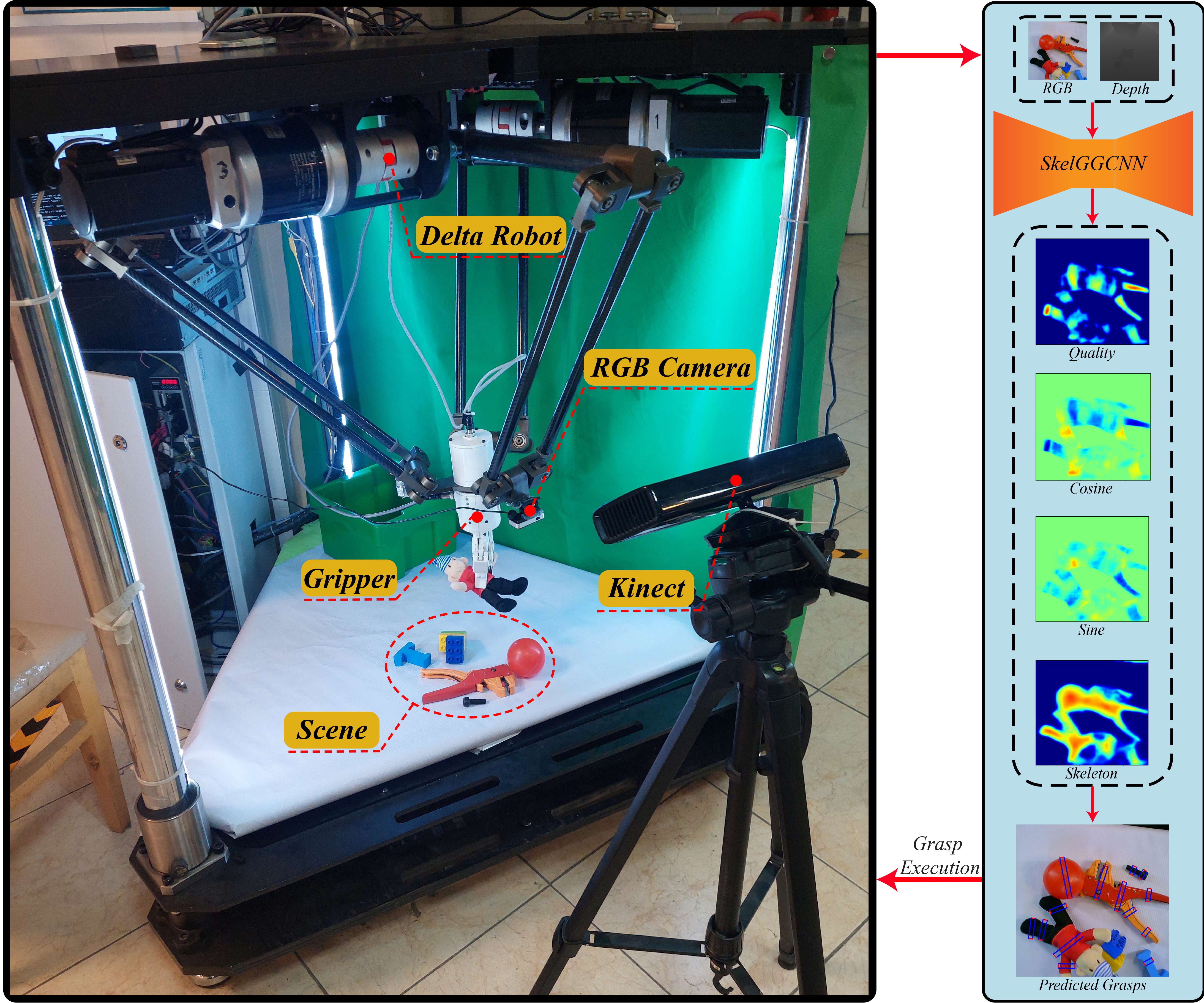}
		\caption{Experimental setup featuring the Delta parallel robot and an illustration of the SkelGG-CNN model’s output for a real-world scene.}
		\label{delta}
	\end{figure}

	This work proposes a comprehensive framework for robotic grasping, integrating a novel grasp pose estimation algorithm, an automated data pipeline, and a 4-DoF learning approach via the Skeleton-Based Generative Grasp Convolutional Neural Network (SkelGG-CNN) model to deliver an efficient and adaptable solution, alongside a generalization of the auto-labeling algorithm for 6-DoF grasp generation.
	The skeleton-based grasp generation algorithm developed in this work is an accurate grasp generation method that eliminates the need for a grasp filtering phase.
	Next, the SkelGG-CNN model is introduced as an innovative enhancement to generative grasping approaches.
	This new architecture addresses several limitations of the widely used existing models, improving performance and robustness in grasp pose prediction, and thereby making the method more effective for real-world applications. 
    Figure \ref{delta} illustrates the model’s output alongside the predicted grasps for a real-world scene, highlighting its practical effectiveness.
	
	The proposed framework is designed as a unified integration of geometry-driven grasp generation and data-driven learning, rather than treating these components as independent contributions.
	Instead of presenting a standalone analytical or purely learning-based method, the framework emphasizes the synergy between automatic data generation and geometry-aware learning.
	In particular, the auto-labeling algorithm serves as the foundation of the learning process by generating structured and physically meaningful grasp annotations, while the SkelGG-CNN model leverages these representations to learn robust 4-DoF grasp prediction strategies.
	This design is motivated by the observation that data availability remains a primary bottleneck in learning-based grasping methods. By embedding geometric priors directly into both the data generation and learning stages, the framework bridges the gap between analytical and data-driven approaches, enabling scalable dataset creation while maintaining compatibility with modern learning paradigms.

\begin{figure*}[tp]
		\centering
		\includegraphics[width=\linewidth]{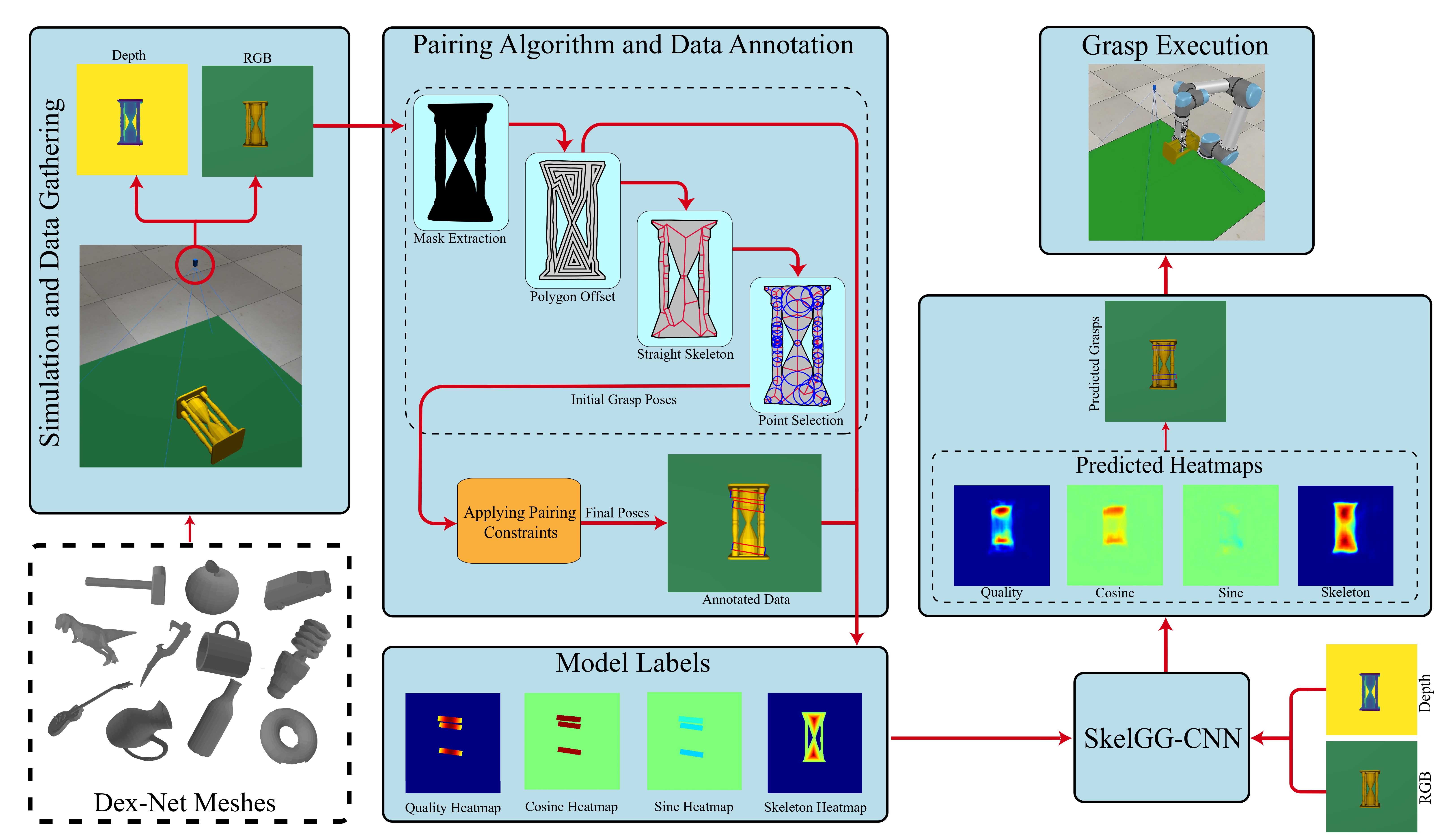}
		\captionsetup{}
		\caption{Overview of the proposed method.
			The pipeline begins with the simulation environment, where RGB-D images are captured from a set of 3D meshes sourced from the Dex-Net dataset. 
			The subsequent stages show the details process of the straight skeleton-based grasp pose estimation algorithm.
			After this process, the generated grasp poses are used to create the SkelGG-CNN model labels.
			The skeleton heatmap is generated during the polygon offsetting phase of the algorithm, adding geometric features to the model.}
		\label{fig:abstract}
	\end{figure*}

	As illustrated in Figure \ref{fig:abstract}, the framework operates as a fully automated pipeline within a simulation environment, enabling large-scale generation of annotated grasp datasets without human intervention.
    The auto-labeling algorithm produces structured grasp annotations that directly support the training of the SkelGG-CNN model, ensuring consistent and scalable dataset creation while maintaining compatibility with modern learning-based methods.

	As a final addition, the auto-labeling pipeline is extended to operate on point cloud data, enabling more comprehensive annotation of 3D grasping scenarios and further enhancing the framework’s applicability to real-world robotic manipulation.
	To process the complex and inherently partial point clouds efficiently, this 6-DoF algorithmic extension employs a generalized quasi-3D approach. Here, quasi-3D refers to a stacked planar strategy that partitions the point cloud into localized horizontal cross sections to apply planar skeletonization.
	The main contributions of this work can be outlined as follows:
	\begin{itemize}
		\item{Introduced a robust and efficient grasp pose estimation algorithm that leverages the geometric skeleton of objects to generate accurate grasp pairs.}
		\item{Developed a fully automated pipeline capable of simultaneously collecting and labeling data without human supervision, effectively addressing the challenge of manual data collection.}
		\item{Proposed SkelGG-CNN, a novel generative model for 4-DoF grasp prediction that utilizes geometric features extracted from the skeleton creation algorithm, enhancing the robustness and quality of predicted grasps compared to state-of-the-art grasping methods.} 
		\item{Trained the model with a significantly smaller dataset, demonstrating the efficiency of geometry-aware learning techniques in learning grasp strategies.}
		\item{Extended the auto-labeling pipeline to work with point clouds using a generalized version of the 2D auto-labeling algorithm, which utilizes a quasi-3D algorithm to generate 6-DoF grasp configurations, showing the capabilities of the skeletonization method in other input modalities.}
		
	\end{itemize}

	This paper extends the skeleton-based grasp keypoint generation method proposed in \cite{b9}.
	The previous work lacked a comprehensive pipeline for grasp annotation and relied on global regions, pairing points from different grasp regions together, which often resulted in failure cases. As a result, the algorithm was computationally expensive and unsuitable for large-scale grasp annotation.
	
	In contrast, the skeleton-based approach presented here has been significantly improved to become a more accurate and efficient algorithm, capable of achieving real-time performance within the automatic pipeline.
	This efficiency arises from restricting the pairing algorithm to local point candidates around each skeleton node rather than considering global regions.
	
	Beyond algorithmic improvements, this work introduces the SkelGG-CNN model, which adds a new output head to traditional generative grasping methods.
	By leveraging skeleton-based grasping regions, the model enables geometry-aware learning and robust generalization.
	This integration of analytical methods with data-driven learning constitutes one of the main contributions of this paper.
	
	Finally, in the mentioned paper, no comprehensive evaluations were conducted, and the experiments were limited to simulation-based grasping tests on a small number of simple objects.
	Moreover, the experiments were constrained to 4-DoF grasping, and the generalization of the auto-labeling algorithm was not discussed in the original paper. 	
	
	Beyond these differences from the earlier work, Table~\ref{tab:comparison} situates the proposed framework among recent grasping approaches by summarizing their dataset-generation capabilities, reported benchmark coverage, cluttered-scene evaluation, and code availability.

	\begin{table*}[tp]
		\centering
		\small
		\renewcommand{\arraystretch}{1.2}
		\setlength{\tabcolsep}{5pt}
		\resizebox{\textwidth}{!}{
			\begin{tabular}{lccccccc}
				\toprule
				\textbf{Author} & \textbf{Automatic Dataset Generation} &\textbf{Cornell Results} & \textbf{Jacquard Results}  & \textbf{GraspNet Results} & \textbf{EGAD! Results} & \textbf{Cluttered Objects} & \textbf{Code Available} \\
				\midrule
				\cite{b69} & \textcolor{red}{$\times$} & \checkmark & \textcolor{red}{$\times$}& \textcolor{red}{$\times$}& \textcolor{red}{$\times$} & \checkmark & \checkmark \\
				\cite{b64} & \textcolor{red}{$\times$} & \checkmark & \checkmark & \textcolor{red}{$\times$}& \textcolor{red}{$\times$} & \checkmark & \textcolor{red}{$\times$} \\
				\cite{b6} & \textcolor{red}{$\times$} & \checkmark & \checkmark & \textcolor{red}{$\times$}& \textcolor{red}{$\times$} & \checkmark & \checkmark  \\
				\cite{b42} & \textcolor{red}{$\times$} & \checkmark & \checkmark & \textcolor{red}{$\times$}& \textcolor{red}{$\times$} & \checkmark &\textcolor{red}{$\times$} \\
				\cite{b92} & \textcolor{red}{$\times$} & \checkmark & \checkmark & \textcolor{red}{$\times$}& \textcolor{red}{$\times$} & \checkmark &\textcolor{red}{$\times$} \\
				\cite{b91} & \textcolor{red}{$\times$} & \checkmark & \checkmark & \textcolor{red}{$\times$}& \textcolor{red}{$\times$} & \textcolor{red}{$\times$} & \textcolor{red}{$\times$}   \\
				\midrule
				This Work & \checkmark & \checkmark & \checkmark & \checkmark & \checkmark & \checkmark & \checkmark  \\
				
				\bottomrule
			\end{tabular}
		}
		\caption{Comparison of various recent grasping approaches across different datasets and approaches.}
		\label{tab:comparison}
	\end{table*}

	\section{Related Work}\label{sec:section2}
	In this section, an overview of different grasping paradigms is provided, with a focus on highlighting the evolution from traditional analytical methods to modern data-driven and geometry-aware approaches.
	Each paradigm is examined in terms of how it addresses the challenges of grasp pose estimation, generalization, and computational efficiency. This discussion provides the foundation for understanding the motivations and contributions of the proposed method.
	\subsection{Analytical Grasping}
	
	Analytical approaches for robot grasping have been extensively studied over the past decades, particularly before neural networks became the dominant paradigm in the field \cite{b20,b27,b29}.
	\cite{b8} addressed the grasping problem by analysing gripper geometry.
	\cite{b61} developed a contact-based grasping approach that adapted to the object’s geometry after initial contact.
	\cite{b25} introduced a geometric approach for suggesting grasps and a method for optimizing them, improving the quality of grasps.
	Modeling human grasping with five fingers is a highly complex task and is often simplified by modeling the grasp as the interaction between two opposing jaws or fingers.
	This abstraction facilitates easier representation and analysis of grasps.
	However, evaluating the quality of grasp poses remains a complex challenge, leading researchers to develop numerous methods and metrics.
	One of the earliest grasp quality metrics, force-closure, evaluates whether all contact forces applied to an object can resist any external wrench exerted on it \cite{b22}.
	\cite{b23} introduced a more efficient algorithm for assessing grasp quality based on force-closure properties. However, the exact calculation of the force-closure metric is often computationally complex and sensitive to noisy inputs.
	Random grasp pose sampling from point clouds and force-closure approximations have been used in the literature to tackle these challenges \cite{b28, b62, b63, b60}.
	The random grasp sampling method can result in many infeasible grasps, and the force-closure metric alone is not sufficient to guarantee grasp success.
	\cite{b9} utilized geometric features of skeletons to reduce the search space and successfully generated highly accurate grasp pairs using specific regions of the object.
	Although analytical methods have become more efficient over the years, analyzing the physical properties of objects remains computationally intensive, making them impractical for the real-time performance demanded by modern robotic systems.
	% \textcolor{red}{$\times$}

	\subsection{Data-Driven Grasping}\label{subsection:22}
	Recent research has shown the effectiveness of deep neural networks, particularly convolutional networks, in learning and capturing relevant features for grasp planning tasks using only RGB-D image inputs, without requiring the full object model \cite{b29}.
	These models often require a large amount of training data.
	The Dex-Net 2.0 dataset \cite{b3} is a large-scale dataset containing thousands of 3D objects. 
	In the original paper, the authors trained a Grasp Quality Convolutional Neural Network (GQCNN) to rank grasp poses, achieving a 93\% success rate on known objects.
	Deep learning approaches for grasp pose generation can be broadly categorized into two main types:
	
	\textbf{1. Direct Grasp Pose Prediction}: Some studies have utilized Convolutional Neural Networks (CNNs) to directly predict the optimal grasp poses \cite{b4,b11}. These methods aim to output the grasp configuration (e.g., position, orientation, and width) directly from the input image.
	
	\textbf{2. Grasp Quality Prediction}: Another widely used architecture for grasping models leverages neural networks to predict grasp quality \cite{b12,b13,b14}. These models typically output a quality score for potential grasp candidates, which can then be used to select the best grasp.
	These data-driven approaches typically require substantial amounts of annotated grasp data, often obtained from large-scale datasets.
	Examples of such datasets include Jacquard \cite{b7}, GraspIt! \cite{b32}, and GraspNet-1Billion \cite{b63}, all of which provide labeled grasp data.
	However, due to the dynamic nature of robotic environments, additional data generation is often required, and the process of creating labeled grasps can be both time-consuming and costly.

\begin{figure*}[tp]
		\centering
		\captionsetup{}
		\includegraphics[width=\linewidth]{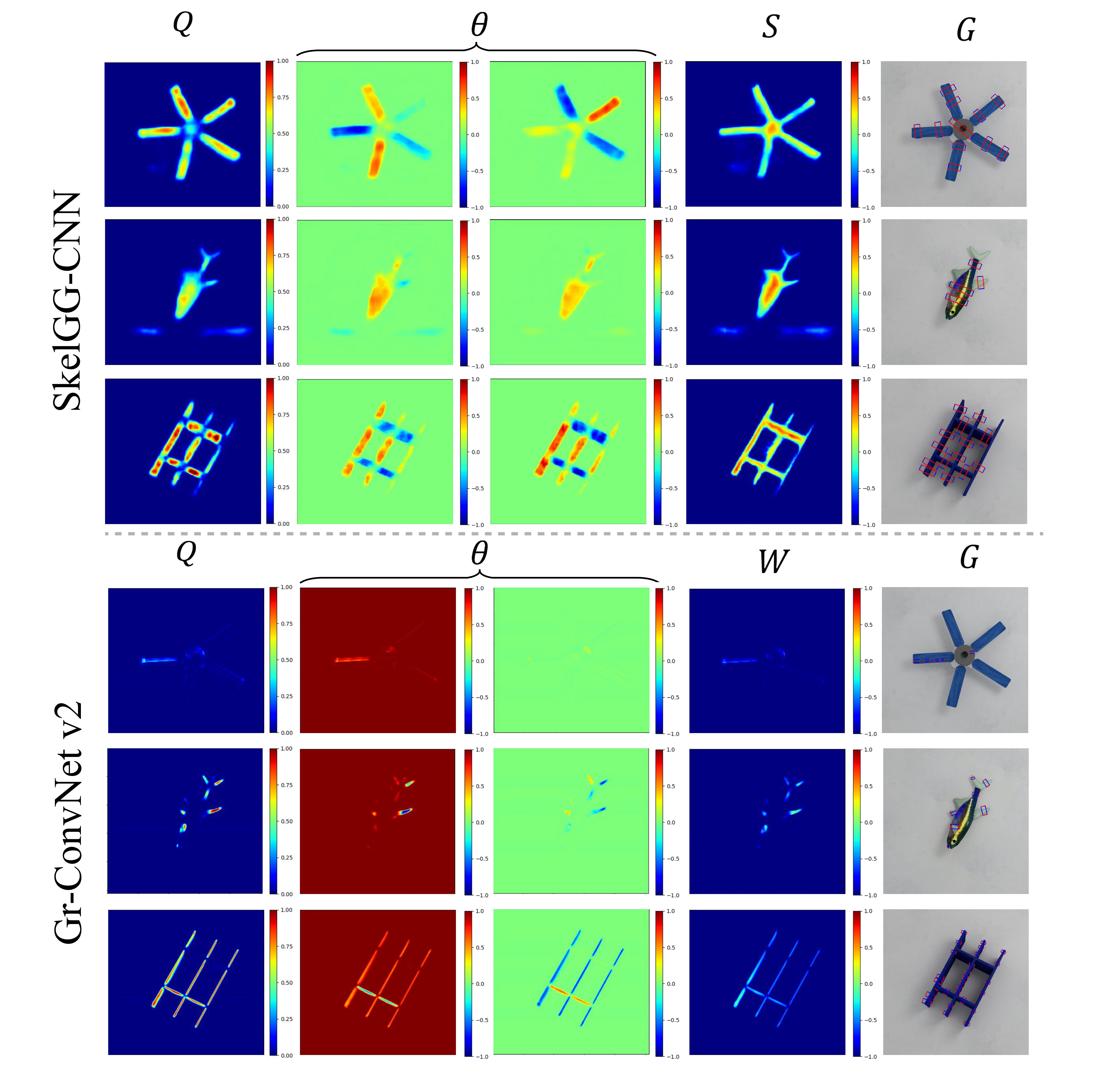}
		\caption{Comparison of SkelGG-CNN output heatmaps with the pretrained Gr-ConvNetV2 \cite{b6} model. The first three columns display the quality and angle heatmaps for both models. The fourth column represents the width heatmap for Gr-ConvNet, whereas SkelGG-CNN uses a skeleton heatmap instead, enabling both grasp width estimation and grasp refinement.}
		\label{fig:jacq_fig_comp}
	\end{figure*}
	
	\subsection{Geometry-Aware Grasp Learning}\label{subsection:23}
	In the modern era, as robotic systems increasingly interact with humans, they should be capable of grasping a wide range of objects.
	This requirement can only be achieved by developing models that can generalize across different objects and environments. 
	A key approach to achieving this is to understand the role of geometric properties of different objects.
	\cite{b15} used primitive shape segmentation to decompose the object point cloud into simpler shapes, thereby teaching the model different primitive geometries.
	Another widely used approach is the idea of pixel-wise or point-wise quality prediction, which has proven highly effective.
	Predicting quality instead of directly generating grasp poses has enabled these models to generalize to cluttered scenes, even when trained solely on single-object images or point clouds \cite{b6,b65,b20,b66,b16}.
	Generative grasp models represent a family of approaches that predict grasp quality.
	These models use RGB-D images of objects to predict pixel-wise heatmaps specifying grasp configurations in the image.
	\cite{b5} used a Generative Grasp Convolutional Network (GG-CNN) to calculate the grasp quality for each pixel.
	Their lightweight generative model takes an RGB-D image as input and generates three heatmaps for grasp quality, angle, and width of the grasp. 
	Using a Generative Residual Convolutional Grasp Neural Network (GR-ConvNet), \cite{b6} achieved higher grasp success rates with a more complex model architecture.
	Their approach achieved 97\% and 94\% accuracy on the Cornell and Jacquard datasets, respectively.
	\cite{b40} used a gradient-based quality heatmap for more accurate grasp center prediction.
	\cite{b41} replaced the width heatmap with a segmentation mask, improving model's width estimation.
	\cite{b42} introduced the HTC-Grasp method, utilizing a transformer-based encoder to capture the global features of the image.
	\cite{b64} introduced four new heatmaps—one for grasp center prediction, and others for grasp orientation offset and grasp keypoint prediction.
	Despite these improvements, determining accurate grasp poses from RGB-D images in real-world environments is still a challenging task.
	A significant limitation is that these models rely on large-scale datasets for training, yet labeled data is often scarce in real-world scenarios.
	The proposed method in this paper aims to address this limitation by learning the geometry of objects, reducing the need for extensive labeled data and enhancing the generalization capabilities of the model.
	
	\begin{figure}[tp]
		\centering
		\includegraphics[width=\linewidth]{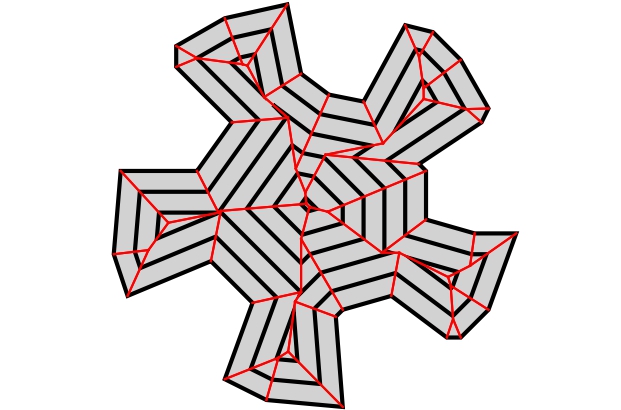}
		\caption{Illustration of polygon offsets with the skeleton specified by red lines.}
		\label{fig:skeleton}
	\end{figure}

	\subsection{Skeletons and Skeleton-Based Grasp Generation Methods}
	Geometric skeleton analysis is a broad area of research, and skeletonization is an important concept, especially in applications where exact analysis of complex object geometries is computationally intensive. 
    Skeletons efficiently represent the geometrical and topological properties of an object, and their applications extend beyond grasping to computer graphics, image processing, and shape analysis, including object recognition, shape matching, and deformation analysis~\cite{b67,b68}.
    By providing a simplified representation of an object's geometry while preserving its key structural properties, skeletons enable efficient analysis across a wide range of applications.
    Various skeletonization methods exist for 2D geometries, with the most notable being the Medial Axis Transform (MAT)~\cite{b82} and the straight skeleton~\cite{b10}. 
    The straight skeleton differs from the medial axis in that it has a straight-line structure and lower combinatorial complexity~\cite{b10}, which can make it preferable in some applications.
    Despite their potential to provide valuable insights into object geometry and grasp pose estimation, applications of skeletons in robotic grasping have been limited.
    \cite{b9} developed a method for grasp keypoint generation using straight skeletons, demonstrating that skeleton structures can effectively reduce the grasp search space by focusing on geometrically meaningful regions of the object.
    However, the method was primarily designed as a standalone analytical grasp generation approach and was not integrated into a large-scale data generation or learning framework.
    \cite{przybylski2011planning} proposed a medial-axis-transform-based grasp planning method using a grid of medial spheres. 
    The method uses local symmetry and object thickness to generate geometrically meaningful grasp candidates and can reduce the search space without relying on simple primitive decompositions. 
    However, it was mainly evaluated in simulation and depends on complete object geometry, such as mesh or point-cloud models, which limits its practicality under noisy and partial real-world perception.
    Moreover, \cite{pennisi2024selective} proposed a learning-based topological skeleton extraction method for selective grasping of complex-shaped industrial parts. 
    Their method uses skeleton keypoints to estimate object pose and select suitable grasping poses, showing the usefulness of skeleton representations in practical robotic manipulation. 
    Nevertheless, the approach is tailored to specific rigid automotive components and relies on predefined skeleton keypoints and object-specific grasping frames, which restricts its generalization to arbitrary unknown objects.

	\subsection{Generative Grasping Convolutional Neural Networks}
    A common architecture for generating grasp poses from RGB-D images is the Generative Grasping Convolutional Neural Networks (GG-CNN) introduced by~\cite{b5}.
    Common generative grasping networks, such as GG-CNN and GR-ConvNet~\cite{b6}, adopt an encoder–decoder architecture that maps an input image $I \in \mathbb{R}^{H \times W \times C}$ (where $C$ is the number of channels, e.g., RGB, RGB-D, or depth) to four prediction heads. 
    Each head produces a heatmap of size $H \times W$, encoding a distinct component of the grasp configuration:
	\begin{itemize}
		\item Quality head $Q \in [0,1]^{H \times W}$: assigns to each pixel $(u,v)$ the probability $Q(u,v)$ of being the center of a successful grasp. This value can also be used as the confidence score of the predicted grasp.
		\item Orientation heads $O_\text{cos}, O_\text{sin} \in [-1,1]^{H \times W}$: predict $\cos(\frac{\theta}{2})$ and $\sin(\frac{\theta}{2})$ for each pixel, which together uniquely define the grasp angle $\theta(u,v) = 2\arctan2(O_\text{sin}(u,v), O_\text{cos}(u,v))$.
		\item Width head $W \in \mathbb{R}^{H \times W}$: specifies the required gripper opening width $w(u,v)$ at each pixel.
	\end{itemize}
	
	At inference time, grasp candidates are generated by first selecting local maxima of the quality map $Q$, and then retrieving their corresponding orientation $\theta(u,v)$ and width $w(u,v)$ from the orientation and width heads. Each grasp candidate is thus parameterized as:
	
	\begin{equation}
		g = (u, v, \theta(u,v), w(u,v))
	\end{equation}
	
	where $g$ represents the final 2D grasp configuration.
	This formulation enables the network to output an effectively unlimited number of grasp poses for a single input image.
	
	Although this architecture has proven effective in real-world applications, it suffers from a key problem.
	The primary issue is that the predicted quality map often identifies multiple, distinct high-quality regions, some of which occur near the object's borders.
	This can lead to inaccuracies in the predicted grasp parameters, such as incorrect orientation or gripper width, ultimately reducing the reliability of the grasp. For instance, Figure~\ref{fig:jacq_fig_comp} illustrates such inaccuracies in the grasp predictions produced by the GR-ConvNet model \cite{b6}. 
    In Section~\ref{sec:section4}, problems with these architectures are addressed, and a new model called the Skeleton-Based Generative Grasp Convolutional Neural Network (SkelGG-CNN) is proposed to overcome these limitations.

	\section{Problem Formulation}
	In this work, the robotic grasping problem is formulated as the predicting of antipodal grasps. 
    A grasp is considered antipodal when the surface normals at the contact points are antiparallel to the finger closure direction and collinear with the line connecting the contact points \cite{b18}.
	It is important to note that while the proposed skeleton-based data generation algorithm is designed to generalize across both planar (4-DoF) and volumetric (6-DoF) spaces, the learning component of the current framework, SkelGG-CNN, is specifically designed to use the 4-DoF representations for high-speed, real-time planar grasp prediction.
	A 4-DoF grasp can be modeled on a 2D plane as a rectangle that specifies the grasp’s position, orientation, and the minimum gripper width required to successfully grasp the object.
	This rectangle can be defined as:
	\begin{equation}
		G_{r} = (P_{c}, \theta, w)
	\end{equation}
	In the above equation, $P_{c} \in \mathbb{R}^2$ is defined as the center of the grasp rectangle, and $\theta$ denotes the grasp’s orientation relative to the horizontal axis.
	The parameter $w$ denotes the minimum gripper width required to successfully grasp the object.
	When projecting to a real-life 4-DoF grasp from the image, the center needs to be defined as $P_{c} \in \mathbb{R}^3$, where the $z$ component represents the depth value corresponding to the $x$ and $y$ coordinates extracted from the depth image.
	Translated to robot coordinates, the grasp can be defined as:
	
	\begin{equation}
		G = (P_{g}, \Theta_{g}, W_{g}) 
	\end{equation}
	
	\noindent where $ P_{g} $ represents the position of the gripper's fingertips, $ \Theta_{g} $ defines the orientation of the gripper, and $ W_{g} $ specifies the distance between the gripper's jaws before grasping.
	The transformation from image coordinates to robot coordinates is performed using a transformation matrix $T$, derived from camera calibration parameters, such that
	\begin{equation}
		G = T(G_{r}) 
	\end{equation}
	This matrix incorporates the camera’s intrinsic and extrinsic parameters, mapping pixel coordinates to the robot’s workspace.
	The \( z \)-component of \( P_g \) (the height of the grasp) is directly inferred from the depth image at the point \( P_c \). By focusing on top-down grasps and assuming a single depth value, the approach avoids the complexities of multi-depth grasp estimation, which can be challenging due to occlusions or irregular object surfaces. This simplification proves effective for tabletop scenarios where objects rest on a flat surface, a common setting in robotic manipulation.
	
	\begin{algorithm}[t]
		\caption{Skeleton-Based Grasp Keypoint Generation Algorithm}\label{alg:alg1}
		\begin{algorithmic}[1]
			\STATE $ \text{\textbf{Input:} Polygon of the shape } \mathbb{P} \text{, straight skeleton of the } $\\
			\hspace{2em} $ \text{polygon } S(\mathbb{P}) \text{, contour points } P $ 
			\STATE $\text{nodes} \gets S(\mathbb{P}). \text{nodes()}$
			\STATE $\text{grasp\_pairs} \gets \{\}$
			\FOR{each node $c_i$ in $\text{nodes}$}
			\STATE $r \gets \text{distance from } c_i \text{ to nearest contour point}$
			\STATE $\text{pair\_candidates} \gets \{P \in C \mid \|P - c_i\| < r + \epsilon \} $
			\FOR{each pair $(P_1, P_2)$ in $\text{pair\_candidates}$}
			\STATE $ \texttt{//Approximate the surface angle} $ \\
			\STATE \hspace{2em}$ \texttt{at $P_1$ and $P_2$} $
			\STATE $ \theta_{1} \gets \text{FitTangentLine}(P_{1}, P)$
			\STATE $ \theta_{2} \gets \text{FitTangentLine}(P_{2}, P)$
			\IF{$ 180^\circ - \epsilon_{\text{angle}} < \angle P_1 c_i P_2 < 180^\circ + \epsilon_{\text{angle}}$}
			\IF{$|\theta_{1} - \theta_{2}| < \epsilon_{\text{angleError}}$}  
			% \small  
			\STATE $\text{grasp\_pairs} \gets \text{grasp\_pairs} \cup \{(P_{1}, P_{2})\}$
			\ENDIF
			\ENDIF
			\ENDFOR
			% \ENDFOR
			\ENDFOR
			\STATE $\textbf{Output:} \text{ grasp\_pairs}$
		\end{algorithmic}
		\label{alg1}
	\end{algorithm}

	\section{Skeleton-Based Grasp Keypoint Generation}
	\label{sec:section4}
	Accurate grasp annotation is a significant challenge in robotic manipulation due to the diversity of object geometries and the labor-intensive nature of manual labeling.
	To overcome the cost and difficulty of manual annotation, this section introduces a novel approach that utilizes the geometric skeleton of objects to develop an efficient algorithm, aiming to bridge the gap between automated and human-level grasp annotation.
	\subsection{Straight Skeleton Construction}
	
	This paper extends the skeleton-based grasp keypoint generation method proposed in \cite{b9}.
	The original work lacked a comprehensive pipeline for grasp annotation and relied on global regions, pairing points from different grasp regions together, which often resulted in failure cases. As a result, the algorithm was computationally expensive and unsuitable for large-scale grasp annotation.
	
	In contrast, the skeleton-based approach presented here has been significantly improved to become a more accurate and efficient algorithm, capable of achieving real-time performance within the automatic pipeline.
	This efficiency arises from restricting the pairing algorithm to local point candidates around each skeleton node rather than considering global regions.

	The straight skeleton of a 2D polygon $P$ is constructed by tracing angular bisectors at each vertex of the polygon.
	As these bisectors are drawn, the polygon progressively shrinks. During this process, two types of events may occur—Edge Events and Split Events—both of which alter the polygon's structure by forming smaller polygons \cite{b10}.
	This shrinking process is recursively applied to the newly formed polygons until the area of the polygons shrink to zero.
	The straight skeleton $S(P)$ can therefore be defined as a graph, whose edges are formed by the angular bisectors. The endpoints of these edges, which are not vertices of 
	$P$, are referred to as nodes.
	Figure~\ref{fig:skeleton} illustrates the skeleton graph of a 2D polygon.
	
	\begin{figure}[tp]
		\centering
		\includegraphics[height=9cm,width=5cm]{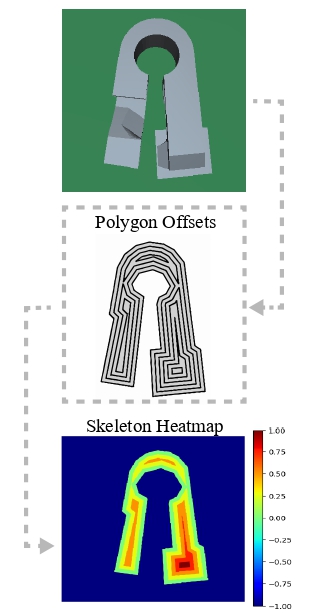}
		\caption{Skeleton heatmap construction steps.}
		\label{fig:gradient}
	\end{figure}

	\subsection{Pairing Algorithm}
	After the skeleton graph construction, the extracted nodes are used to generate grasp keypoints. 
	For each node in the skeleton graph, a circle is inscribed within the polygon, centered at the node, with an initial radius equal to the distance to the nearest contour point.
	This radius is slightly increased to ensure that more points from the object's contour are sampled for keypoint pairing.
	The selected points on each circle are paired based on specific constraints to identify pairs that satisfy the property of being antipodal grasps.
	To apply these constraints, the vertex angles corresponding to a given pair of points on the polygon are computed.
	These angles should correspond to lines that are approximately parallel and tangent to the line connecting the points.
	To take the physical volume of the gripper into account and to ensure that it can approach and grasp the object without collision, three circles are constructed along the line connecting the grasp pairs, extending outward from both sides.
	The radii of these circles are chosen to represent the space required by the gripper during grasping.
	The intersections of these circles with the object's polygon are then measured.
	If any overlap is detected, the grasp pair is considered invalid, as the gripper would collide with the object.
	This ensures that only grasp pairs with sufficient clearance for the gripper are selected.
	The pseudocode in Algorithm~\ref{alg:alg1} provides a detailed explanation of the pairing algorithm.
	
	The developed pairing algorithm efficiently generates multiple grasp pairs by leveraging geometric constraints and localized searches, thereby avoiding an exhaustive exploration of all potential points on the object's surface. 
	This targeted approach significantly reduces computational overhead while maintaining high accuracy in identifying feasible grasp pairs, making the method both time-efficient and practical for large-scale grasp data annotation.
	This efficiency arises from the small number of skeleton graph nodes and the algorithm only searches for the nearest contour points to the nodes, and selects grasp pairs in a greedy manner.
	Straight skeletonization naturally decomposes complex shapes into local geometric regions, and these local regions provide useful grasp candidates.
	Geometric insights from straight skeletons are used to address the labor-intensive challenge of manual labeling, offering a scalable solution for training the SkelGG-CNN model on diverse object geometries.

	\begin{algorithm}[t]
		\caption{Grasp Width Estimation and Refinement Algorithm}\label{alg:alg2}
		\begin{algorithmic}[1]
			\small
			\STATE $ \text{\textbf{Input:} Model Predicted Quality Heatmaps} H_{Q} \text{, } $
			\hspace{2em} $\text{Orientation Heatmap } H_{\theta} \text{, and Skeleton Heatmap } H_{S}  $\\
			\STATE $P_{c_{i}} \gets \text{GetHighestQualityCenters}(H_{Q}) $
			\STATE $ \theta_{i} \gets H_{\theta}(P_{c_{i}})$
			\STATE $ \texttt{//Grasp refinement steps}$
			\STATE $P_{c} \gets \text{argmax}\big(H_{S} \text{ along the line from } P_{c_{i}} \text{ in direction } \theta_{i}\big)$
			\STATE $ \theta \gets H_{\theta}(P_{c})$
			\STATE $w \gets \text{CalculateWidth}(P_{c}, \theta, H_{S})$
			\STATE \textbf{Output:} $(P_c, \theta, w) $
		\end{algorithmic}
	\end{algorithm}

	\subsection{Enhancing Skeleton-Based Grasping Using Model-Based Prediction}
	The developed auto-labeling algorithm generates high-quality grasp keypoints for a wide range of objects and geometries.
    However, when applied to cluttered scenes with multiple objects, the method faces limitations. 
    Specifically, it struggles to distinguish between object boundaries, often producing grasp candidates that lie between two adjacent objects.
	
	To address this challenge, a more intelligent strategy is required.
	Incorporating a neural network provides a promising solution, as it can learn robust grasping strategies for single objects and generalize them to cluttered scenes. 
	In this context, the auto-labeling algorithm serves as a valuable source of training data, while the neural network builds upon it to adapt to more complex scenes.
	
	Most importantly, the proposed neural network is specifically designed to learn geometry-aware grasps. 
	Additionally, the generative grasp model employed in this work assigns confidence values to its predictions. 
	These confidence scores make it possible to filter out low-quality grasps, resulting in more reliable and effective grasping across diverse environments.
	These advantages are the main reason for using a hybrid approach to obtain the benefits of both methods.
	The following sections demonstrate that the SkelGG-CNN model can also outperform the auto-labeling method, as evidenced by higher success rates and accuracies (detailed in Section~\ref{sec:section5}).
	
	\begin{figure}[tp]
		\centering
		% \captionsetup{justification=centering}
		\includegraphics[width=\linewidth]{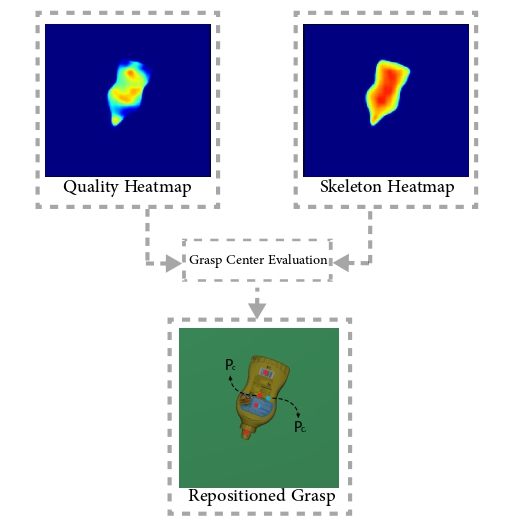}
		\caption{Example of grasp repositioning, the initial grasp center \( P_{c_i} \) from the quality heatmap is shifted along the orientation \( \theta_i \) to the point \( P_c \) with the maximum skeleton heatmap value. This refinement produces more robust grasps compared to the approaches that only use quality maps.
			In this figure, blue dots represent initial grasp centers and red dots indicate the final grasp centers after the refinement process.}
		\label{fig:skelqualrefine}
	\end{figure}

	\begin{figure*}[tp]
		\centering
		\includegraphics[width=\linewidth]{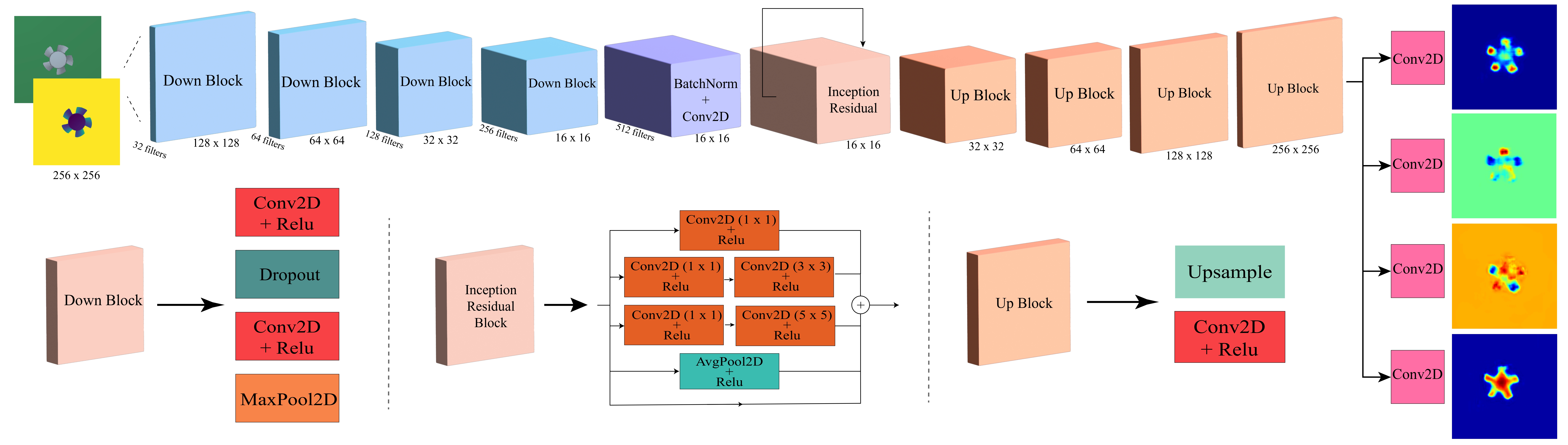}
		\caption{Architecture of the SkelGG-CNN model.}
		\label{fig:model1}
	\end{figure*}
	
	\subsection{SkelGG-CNN Grasp Prediction and Refinement with Skeleton Heatmap}

	The architecture of the SkelGG-CNN model includes four output heads, each contributing to the formation of the grasp configuration. 
	The first three heads generate the grasp center ($P_{c}$) and orientation ($\theta$).
	In contrast to existing generative methods, the label for the fourth head is constructed using the sequence of offset polygons generated during the skeleton construction of the auto-labeling process.
	The skeleton heatmap is created by assigning a value of $-1$ to all pixels outside the object's polygon.
	Inside the polygon, pixel values increase progressively from the outer edges toward the offset polygons.
	These offset polygons shrink toward the local centers of the object's geometry, resulting in those areas having the highest values.
	This happens because during the offsetting process, multiple polygons may be created, producing multiple local centers throughout the object.
	Figure~\ref{fig:gradient} illustrates the construction of the skeleton heatmap.
	When calculating the gripper's opening width, a line is drawn from the grasp center along the specified orientation.
	The width is determined by extending the line until it intersects the polygon's edges, at which point the values become negative.
	The length of this line segment corresponds to the gripper's required opening width.
	Additionally, this heatmap enables the model to refine the grasp configuration.
	The refinement process works as follows: after the initial grasp center is determined using the quality heatmap and an initial grasp orientation is identified, the grasp is adjusted along this orientation toward the point with the highest value in the skeleton heatmap.
	Algorithm~\ref{alg:alg2} details the grasp width calculation and grasp refinement process.

	This refinement offers multiple advantages.
	First, it enhances the ability to reposition the grasp, thereby improving the accuracy of the predicted grasp points, as demonstrated in Figure~\ref{fig:skelqualrefine}, where the grasp is adjusted to better align with the object's geometry.
	By providing a more reliable framework for grasp adjustment, the system becomes more effective in real-world applications.
    Second, during the development process, it was observed that learning to generate this specific heatmap particularly facilitates learning of the quality heatmap, leading to more robust heatmap predictions.
	
	As illustrated in Figures~\ref{fig:skelqualrefine} and~\ref{fig:jacq_fig_comp}, the skeleton heatmap activations consistently align with the high-confidence regions of the quality map. 
    This correlation is further supported quantitatively by the ablation study in Section~\ref{sec:section5} (Figure~\ref{fig:ablation_plot}), 
    which shows that the skeleton map head of SkelGG-CNN achieves both faster convergence and higher final accuracy compared to the width map head, while both variants substantially outperform the GG-CNN~2 baseline. 
    These results suggest that skeleton heatmaps serve as a rich source of contextual information, enabling the model to better predict the quality and feasibility of a given grasp.

\begin{figure*}[tp]
		\centering
		\captionsetup{}
		\includegraphics[width=\linewidth]{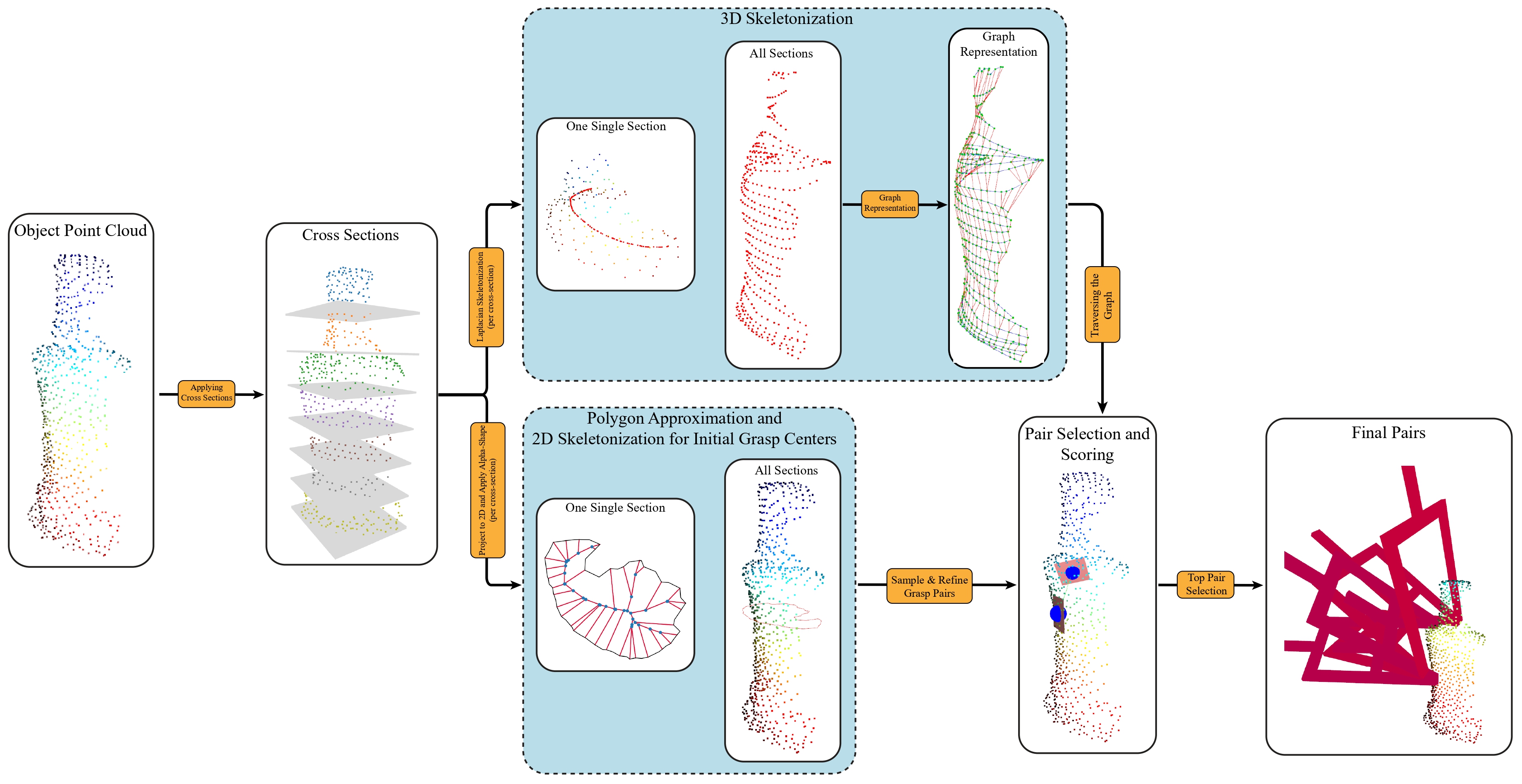}
		\caption{Overview of the generalized 6-DoF skeleton-based auto-labeling pipeline. (Left) The input object point cloud is partitioned into overlapping horizontal cross-sections (80\% overlap) orthogonal to the $z$-axis. (Center — upper path: 3D skeletonization) Each slice is processed with a Laplacian contraction skeletonization; skeleton points within a slice are connected using a Minimum Spanning Tree (MST) and then linked to nearest neighbors in adjacent slices to produce a compact 3D skeleton graph that captures the object’s volumetric structure. (Center — lower path: 2D polygon approximation and initial centers) Each cross-section is projected to 2D and approximated with an alpha-shape polygon; a straight skeleton computed on this polygon yields initial central grasp centers. (Right) Candidate skeleton-node pairs are sampled and refined by traversing the 3D skeleton (depth-first search) from these initial centers; at each candidate pair a local plane is fitted to neighboring skeleton edges and the antipodality of the planes is evaluated as the grasp quality measure. (Far right) Top-scoring pairs are selected and visualized as final 6-DoF gripper poses superimposed on the original point cloud.}
		\label{fig:6dof}
	\end{figure*}

	\subsection{SkelGG-CNN Architecture}
    The SkelGG-CNN is a fully convolutional encoder-decoder network designed for grasp pose estimation, producing pixel-wise heatmaps that capture grasp position, orientation, and geometric refinement. 
    The overall structure of the proposed architecture is shown in Figure~\ref{fig:model1}. 

    In the downsampling stage, features are compressed through four convolutional blocks with increasing numbers of filters (32--256), ReLU activations, dropout, and max-pooling.
    Dropout is applied in the downsampling stage to reduce overfitting, which is particularly important given the relatively small size of the Dex-Net training set used in this work.

    Inception-style multi-scale convolutions were selected because they capture features at different receptive field sizes simultaneously~\cite{szegedy2015} and the residual connection helps with the convergence of the model.
    Batch normalization is applied at the bottleneck to stabilize training at this high-channel-count stage.
    The bottleneck expands features to 2048 channels before reducing them back to 512, allowing the module to capture rich multi-scale representations without propagating excessive dimensionality to the upsampling path.

    The upsampling path mirrors the encoder with nearest-neighbor upsampling and convolutions, reducing filters back to 32. The combination of upsampling and convolution was preferred over transposed convolutions, which, although widely used for feature map upscaling, were found to introduce training instabilities such as inconsistent convergence and checkerboard artifacts, ultimately degrading overall model performance.
    Finally, from the final feature maps, four output heads generate heatmaps for grasp quality, orientation (via sine and cosine), and skeleton geometry.
	
	\subsection{Training}
	The model is trained using a smooth L1 loss function, which some prior generative grasping works found effective for handling exploding gradients during training~\cite{b6, b64}.
	The overall loss is defined as the sum of the individual loss components:
	\begin{equation} 
		\mathcal{L} = \mathcal{L}_{Q} + \mathcal{L}_{\theta} + \mathcal{L}_{S},
	\end{equation}
	where $\mathcal{L}_{Q}$, $\mathcal{L}_{\theta}$, and $\mathcal{L}_{S}$ represent the loss terms associated with the quality, angle, and skeleton heatmaps, respectively. The function $\mathcal{L}$ for each loss term is calculated as:
	\begin{equation}
		\mathcal{L}(y,\hat{y}) = \frac{1}{N}\sum_{i=1}^{N} \mathrm{SmoothL1}(y_{i} - \hat{y}_{i})
	\end{equation}
	
	where the Smooth L1 is defined as:
	\begin{equation}
		\text{SmoothL1}(x) =   
		\begin{cases}   
			0.5x^2 & \text{if } |x| < 1 \\   
			|x| - 0.5 & \text{otherwise}   
		\end{cases}  
	\end{equation}  
	Here, $ y $ and $\hat{y} $ are the labels and the predicted outputs, respectively.
	The Adam optimizer was employed with an initial learning rate of \(10^{-3}\), which was reduced by a factor of 0.1 every 25 epochs to ensure stable convergence.

    \subsection{Generalized 6-DoF Skeleton-Based Auto-Labeling}
    \label{subsec:6dof_method}

    The skeleton-based auto-labeling algorithm presented earlier in this section is limited to generating 4-DoF grasp annotations.
    Despite this limitation, the 4-DoF grasps remain widely used in modern robotic applications due to their computational efficiency, compatibility with common 4-DoF robotic hardware (e.g., Delta and SCARA robots), and suitability for structured environments like tabletop or conveyor-based pick-and-place tasks \cite{noh2025graspsam, hu2025push}. 
	Their simplicity enables robust, high-speed grasping with minimal sensitivity to noisy depth data, making them prevalent in industrial settings despite advancements in 6-DoF grasping.
	Recent research, as discussed in Section~\ref{sec:section2}, validates these advantages through data-driven 4-DoF grasping models, which achieve high accuracy in structured scenarios.
	Nevertheless, this restriction often becomes inadequate in highly cluttered scenes, where top-down approaches may not provide feasible grasping solutions, and the robot requires more flexible orientations to reach and manipulate objects successfully.

	The straight skeleton approach uses the geometric features of 2D polygons 
	to find local grasping regions in objects.
	Unfortunately, this technique does not extend naturally to 3D point clouds and meshses, as 3D meshes lack the same well-defined polygonal properties.
	While several skeletonization algorithms have been proposed for 3D meshes and point clouds \cite{wen2024glskeleton,cao2010point,jalba2012surface}, they typically assume access to complete mesh or point cloud information.
	In real-world robotic setups, however, point clouds are often noisy, partial, and incomplete, leading these methods to produce inaccurate or unstable skeletons.
	To address the discussed issue and utilize the previously developed 2D skeletonization approach, the following quasi-3D approach was developed. 
	By using this quasi-3D formulation, the developed algorithm aims to preserve essential 3D topological features while remaining robust to incomplete, noisy depth data.

	The proposed generalized skeleton-based auto-labeling algorithm begins by partitioning the input point cloud into horizontal cross-sections orthogonal to the $z$-axis.
	Within each cross-section, a Laplacian-based skeletonization algorithm~\cite{cao2010point} is applied.
	This iterative contraction method progressively reduces the point set into a one-dimensional skeleton by moving points toward the local medial regions of the surface.
	The transformation at each iteration is expressed as:

	\begin{equation} 
		\begin{bmatrix} W_L L \\ W_H 
		\end{bmatrix} P' = 
		\begin{bmatrix} 0 \\ W_H P 
		\end{bmatrix} 
	\end{equation}
	
	where $L \in \mathbb{R}^{n \times n}$ as the paper denotes, is the cotangent-weight Laplacian matrix, $W_L$ and $W_H$ are weighting diagonal matrices for controlling the contraction, $P \in \mathbb{R}^{n \times 3}$ is the original point set, and $P' \in \mathbb{R}^{n \times 3}$ is the contracted point set.
	When applied to horizontal cross-sections that do not form closed surfaces, the skeletonization procedure produces skeleton curves that follow the boundary of the mesh and capture the outer volume of the object. 
	Repeating this process across all cross-sections reconstructs the overall object volume. 
	To account for topological variations between successive slices, each skeleton point in one cross-section is connected to its nearest skeleton point in the adjacent upper layer. 
	Within each layer, the points are further connected using the Minimum Spanning Tree (MST) algorithm.
	Cross sections are sampled with an 80\% overlap, ensuring smooth transitions and continuity throughout the reconstructed volume.
	
	\begin{table}[t]
		\centering
		\caption{Performance comparison of algorithms on Accuracy with IoU (\%) on Jacquard V1 dataset.}
		\label{tab:rgbd_comp}
		\begin{tabular}{@{} l c c@{}}
			\toprule
			& \textbf{Input type} & \textbf{Jacquard accuracy(IoU)} \\ \midrule
			& RGB & 94.46 \\
			& Depth & 95.32 \\
			& RGB-D & 95.42 \\
			\bottomrule
		\end{tabular}
		
	\end{table}

	The final result is a skeleton graph that compactly represents the structure of the object. As the skeleton cross-sections have exactly the same number of points, the final skeleton is heavily downsampled for more efficient representation.
	This skeleton graph serves as a foundation for sampling candidate grasp regions and the following pair selection.
	
	After constructing the skeleton graph, a set of initial grasp centers is required to initiate the pairing process. 
	To generate these centers, each horizontal cross-section is projected into 2D space.
	A polygonal approximation of the cross-section boundary is then computed using the alpha-shape algorithm, which yields a 2D polygonal representation of the cut.
	From this polygon, the straight skeleton algorithm is applied to extract central points, which serve as the initial grasp centers.
	These centers act only as initial points for the subsequent refinement stage.
	
	In the refinement step, pairs of nodes from the 3D skeleton graph are sampled, and the graph is traversed to search for pairs corresponding to optimal grasp configurations.
	The traversal is performed using a depth-first search (DFS), rooted at the initially sampled nodes.
	At each candidate pair, a local plane is fitted to the neighboring edges of the skeleton at that node, and the alignment of these planes is evaluated as a grasp quality measure.	
	
	To improve efficiency, the traversal avoids revisiting ancestor nodes once a better candidate pair with more opposing normals has been identified.
	The proposed skeletonization procedure demonstrates strong robustness to noisy and incomplete point clouds, which are commonly encountered in real-world robotic setups and datasets.
	Operating on localized cross-sectional structures eliminates the need for a fully closed surface representation and reduces the instabilities typically associated with isotropic 3D skeletonization methods.
	As a result, the method produces stable and consistent skeletal representations even under significant occlusions or missing data, leading to improved 3D shape approximation compared to approaches that rely on complete volumetric information.
	In contrast to fully isotropic 3D skeletonization methods, which often degrade under partial observations, the proposed formulation maintains stability by relying on locally consistent geometric structures.
	The entire pipeline of the proposed generalized skeleton-based 6-DoF auto-labeling algorithm is illustrated in Figure~\ref{fig:6dof}.
	
	The use of horizontal cross-sections reflects a deliberate trade-off between geometric completeness and computational efficiency.
	While isotropic 3D slicing could theoretically capture all volumetric directions, it would introduce substantial computational overhead and generate a large number of redundant or physically infeasible grasp candidates.
	Restricting the analysis to horizontal slices instead provides a structured and efficient decomposition of the geometry, aligning well with common tabletop setups and object configurations.
	Empirical evaluation shows that this strategy yields a diverse and sufficient set of grasp configurations across a wide range of object shapes and orientations.
	Furthermore, the subsequent refinement stage operates directly on the 3D skeleton graph and leverages local geometric information, compensating for the reduced directional sampling introduced during the slicing stage.

	\section{Experiments and Results}
	\label{sec:section5}
	This section presents the results of various experiments and introduces evaluation strategies.
	The main purpose of these experiments is to demonstrate the capability of the skeleton-based grasp generation approach as an effective annotation method.
	Furthermore, it is important to show that the model trained with the generated labels, which in this work is SkelGG-CNN, has the ability to generalize based on these grasps to be a geometry-aware model that can perform well across different benchmarks and experimental environments.

	\begin{table}[t]
		\centering
		\small
		\small
		\caption{Comparison of model performance with different parameter count on the Jacquard V1 dataset.}
		\label{tab:size_cmp}
		\begin{tabular}{p{3.2cm} p{2.3cm} p{1.7cm}}
			\toprule
			\textbf{Model parameters} & \textbf{Accuracy (IoU)} \\
			\midrule
			951,316 & 92.10 \\
			3,801,252 & 95.11 \\
			15,196,996 & 95.42 \\
			\bottomrule
		\end{tabular}
	\end{table}

	\subsection{Datasets}
	Several well-known datasets are used in this work for both training and benchmarking, each offering unique characteristics in terms of annotation methods and object diversity.
	
	\textbf{Jacquard V1:} The Jacquard dataset \cite{b7} is a large-scale grasping dataset containing 54k RGB-D images of single objects from a diverse range of items gathered in a simulation environment, comprising 1,181,330 annotated grasps. The annotations are generated using automated sampling methods, and evaluation is performed with a grasp accuracy metric based on the Intersection over Union (IoU):
	\begin{equation}
		\mathrm{IOU}(\overline{R}, R) = \frac{|\overline{R} \cap R|}{|\overline{R} \cup R|}
	\end{equation}
	A grasp is considered successful if the IoU is greater than 0.25 and the orientation error relative to the ground-truth is less than $\frac{\pi}{6}$.
	
	\textbf{Jacquard V2:} A refined version of the original dataset, Jacquard V2 \cite{b75} applies a Human-In-The-Loop approach to enhance annotation quality. While further details are provided in Section~\ref{subsec:eval_jacquardV2}, it is important to note that this refinement aligns the automated labels more closely with human-level annotations.
	
	\textbf{Cornell Dataset:} This dataset is a widely used benchmark in robotic grasping, comprising 885 RGB-D images of 240 distinct objects. Grasp annotations in this dataset are manually labeled by human annotators, ensuring higher accuracy. Consequently, evaluations conducted on this dataset are considered more reliable due to the precision of its annotations.
	
	\textbf{Dex-Net Dataset:} The Dex-Net dataset contains 3D meshes of a diverse range of objects. In this study, 437 unique objects were selected to form the training dataset for the SkelGG-CNN model, resulting in 2,622 images and 35,322 annotated grasps. Additionally, a test
	dataset comprising 125 household object meshes was created for evaluation purposes. 
	The training and test datasets consist of a variety of geometries, including objects such as fruits, cups, bottles, toys, cans, and tools.
	Finally, to evaluate the model’s generalization capabilities, a separate set of 25 complex, unseen objects was used for further evaluations.
	
	\textbf{GraspNet-1Billion Dataset:} The GraspNet dataset consists of 190 cluttered tabletop scenes captured under realistic settings with different difficulties, where multiple household objects are placed together in diverse configurations~\cite{b63}. 
	The scenes are recorded from different viewpoints and under varying levels of occlusion, providing a challenging environment for evaluating robotic grasping performance. 
	
	\begin{table}[t]
		\centering
		\scriptsize
		\caption{Performance comparison of algorithms on Accuracy with IoU (\%) on the Jacquard V1 dataset.}
		\label{tab:jacq_comp}
		\begin{tabular}{p{2.15cm} p{2.06cm} p{0.9cm} p{0.6cm} p{0.6cm}}
			\toprule
			\textbf{Authors} & \textbf{Algorithm} & \textbf{Accuracy} & \multicolumn{2}{c}{\textbf{Labels}} \\
			\cmidrule(lr){4-5}
			\textbf{} & \textbf{} & \textbf{} & \textbf{Train} & \textbf{Test} \\
			\midrule
			\cite{b72} & FCGN, ResNet101 & 92.80 & GT & GT \\
			\cite{b73} & ROI-GD, ResNet101 & 93.60 & GT & GT \\
			\cite{b5} & GG-CNN 2 & 84.00 & GT & GT \\
			\cite{b71} & DD-Net, Hourglass & 97.00 & GT & GT \\
			\cite{b74} & SE-ResUNet & 95.70 & GT & GT \\ 
			\cite{b6} & GR-ConvNet & 95.10 & GT & GT \\ 
			\cite{b42} & HTC-Grasp & 95.80 & GT & GT \\
			\cite{b91} & Grasp-DSC & 92.20 & GT & GT \\
			\midrule
			This Work & Auto-Labeling & 89.20  \\
			This Work & SkelGG-CNN (RGB-D) & 95.42 & StSkel & GT \\
			\bottomrule
		\end{tabular}
	\end{table}

	\textbf{EGAD! Dataset:}
	The EGAD! dataset~\cite{morrison2020egad} is designed to evaluate robotic grasping across objects of varying levels of difficulty.
	It contains a diverse set of geometrically complex and simple objects, allowing for systematic testing of how grasping algorithms generalize to unseen shapes.
	By measuring the grasp success rate on objects with different complexity levels, the dataset provides valuable insights into the robustness and adaptability of grasp generation methods.

	\begin{table}[t]
		\centering
		\scriptsize
		\caption{Performance comparison of algorithms on Accuracy with IoU (\%) on the Jacquard V2 dataset.}
		\label{tab:jacquard_V2}
		\begin{tabular}{p{2.15cm} p{2.05cm} p{0.9cm} p{0.6cm} p{0.6cm}}
			\toprule
			\textbf{Authors} & \textbf{Algorithm} & \textbf{Accuracy} & \multicolumn{2}{c}{\textbf{Labels}} \\
			\cmidrule(lr){4-5}
			\textbf{} & \textbf{} & \textbf{} & \textbf{Train} & \textbf{Test} \\
			\midrule
			\cite{b76} & ResNet50 & 91.37 &  GT & GT \\
			\cite{b76} & ResNet101 & 91.35 & GT & GT \\
			\cite{b76} & ResNet152 & 90.30 &  GT & GT \\
			\cite{b77} & Squeeze & 94.76 &  GT & GT \\
			\cite{b78} & ShuffleNet & 94.85 &  GT & GT \\
			\cite{b79} & Xception & 95.27 &  GT & GT \\
			\cite{b69} & GG-CNN & 94.54 &  GT & GT \\
			\cite{b5} & GG-CNN 2 & 97.24 &  GT & GT \\
			\cite{b80} & MobileNetV2 & 93.05 &  GT & GT \\
			\cite{b80} & MobileNetV2Pruning & 96.12 &  GT & GT \\
			\midrule
			This Work & SkelGG-CNN & 96.55 & StSkel & GT \\
			\bottomrule
		\end{tabular}
	\end{table}

	\subsection{Evaluation Strategy}
	In order to evaluate the effectiveness of both the auto-labeling algorithms and the trained model, five distinct experiments were designed, each targeting specific aspects of the framework’s performance: its capacity as a fully automated pipeline, the validity of its auto-generated labels, and its generalization across different data distributions.
	Each dataset used in this section provides unique evaluation perspectives for the framework.
	For the Jacquard dataset, its RGB-D images are processed, and the auto-labeling algorithm generates grasp labels, enabling assessment of annotation robustness across diverse object categories.
	For the Cornell dataset, no new labels are produced with the algorithm; instead, its human-annotated ground-truth labels are used exclusively for testing to evaluate cross-dataset generalization. 
	
	The Jacquard and Cornell datasets are suitable for 4-DoF grasping evaluation.
	However, to address their limitations and allow for full 6-DoF benchmarking, three additional datasets are used.
	For these evaluations, an improved and generalized version of the auto-labeling algorithm for 6-DoF grasp generation is introduced, as described in Section \ref{subsec:6dof_method}.
	
	First, the performance of the generalized skeleton-based auto-labeling algorithm is measured on the GraspNet dataset.
	The cluttered scenes and complex point clouds in GraspNet make it an ideal choice for evaluating 6-DoF grasping methods.
	To further test grasp success, experiments are carried out on the EGAD! dataset. The 4-DoF SkelGG-CNN model (trained via the 2D auto-labeler) is evaluated to assess performance under severe depth ambiguities, and the 6-DoF geometric algorithm is evaluated separately to demonstrate how full 3D information overcomes these limitations.
	The variety and shape complexity in EGAD! highlight the limitations of 2D grasping methods and demonstrate the generalized algorithm’s ability to address them.
	
	Finally, the Dex-Net dataset is used to evaluate and compare all developed methods on common household objects as well as more complex geometries.
	This combination of datasets and labeling methods forms the basis of the evaluation strategy described below.
	
	\begin{figure}[tp]
		\centering
		\captionsetup{}
		\includegraphics[width=0.8\linewidth]{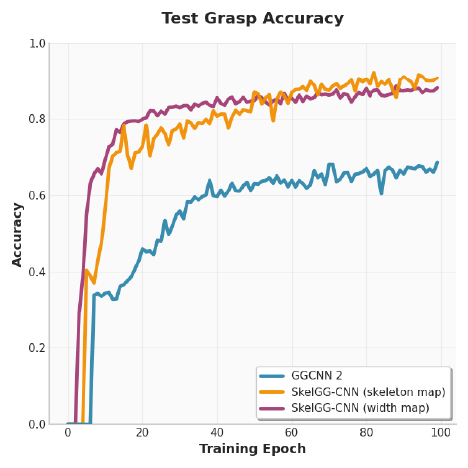}
		\caption{Comparison of test accuracies of the trained models on Dex-Net dataset.}
		\label{fig:ablation_plot}
	\end{figure}
	
	\textbf{Evaluation on Jacquard with Ground-Truth Testing:}\\
	In this evaluation, images from the Jacquard V1 and Jacquard V2 datasets are first processed and auto-labeled using the developed algorithm.
	The SkelGG-CNN model is then trained on these auto-generated labels, using 80\% of the dataset for training.
	Its performance is then evaluated on the separate test sets of Jacquard V1 and V2, with accuracy determined using the datasets's original ground-truth labels.
	This experiment seeks to demonstrate the high accuracy of the auto-labeling method on a large-scale dataset, which offers a standardized framework for comparing various approaches.

	\textbf{Cross-Dataset Generalization on Cornell:}\\
	In this evaluation, the performance of the SkelGG-CNN model is measured on a different dataset on which the model has not been trained.  
	For this experiment, real images from the Cornell dataset are given to the model, with accuracy calculated using Cornell’s manually annotated ground-truth labels.
	Comparing the results with human-annotated labels demonstrates the robustness of both the auto-labeling algorithm and the SkelGG-CNN model.
	Strong performance in this context highlights the framework’s adaptability to varied real-world scenarios without requiring dataset-specific retraining or labeling.

	\textbf{Evaluation on GraspNet Benchmark With a Generalized Auto-Labeling Approach for 6-DoF Grasp Generation:}\\
	The GraspNet dataset is mainly designed as a benchmark for training and evaluating model-based approaches in 6-DoF grasp pose estimation.
	It uses a force-closure metric with different friction coefficients to assess grasp quality.
	Since the developed method was initially designed for 4-DoF grasping, it could not be directly applied to this benchmark. 
	To overcome this, a generalized auto-labeling strategy was introduced to extend the method to 6-DoF, enabling the generation of high-quality grasp poses on point clouds.
	
	\textbf{Evaluation on EGAD! for Measuring Limitations of 4-DoF Grasping Methods:}\\
	To provide a comprehensive assessment of performance, the EGAD! benchmark is utilized, as it is specifically designed to challenge the limits of depth perception and geometric understanding.
	As stated by the authors of the EGAD! paper, the complexity of this benchmark exposes the limitations of traditional 2D-based approaches, particularly in depth estimation.
	For this evaluation, 49 objects from the dataset’s evaluation set were simulated.
	The evaluation mainly shows that the generalized auto-labeling approach effectively overcomes the shortcomings of 2D methods and performs reliably even under the most challenging geometric conditions.
	
	\textbf{End-to-End Framework Evaluation on Dex-Net (Simulation Experiments):}\\
	Dex-Net’s CAD models are collected, simulated, and auto-labeled through an automated pipeline.
	The SkelGG-CNN model is then trained on these auto-generated labels.
	The trained model is subsequently evaluated on both a test set and a set of novel, unseen objects within the simulation environment.
	Success rate is measured by the robot’s ability to grasp and hold an object for 3 seconds.
	This process illustrates the framework’s capability to function as a fully automated pipeline—covering dataset collection, auto-labeling, model training, and evaluation—all within the same simulation environment.
	Moreover, the pipeline generalizes well from simulation to real-world systems, as will be demonstrated in Section \ref{sec:section6}.
	Finally, the 6-DoF auto-labeling is also evaluated in the same simulation setup as a tool to compare the generalized algorithm's performance on challenging geometries with the original algorithm's performance.
	
	These evaluations collectively demonstrate the framework’s performance. Dex-Net, Jacquard, and Cornell datasets validate the 2D (4-DoF) end-to-end pipeline, auto-generated labels against a benchmark, and generalization to unseen data, respectively. GraspNet assesses 6-DoF grasp generation, while EGAD! serves a dual purpose: exposing the geometric limits of 4-DoF learning models and demonstrating the robustness of the proposed 6-DoF geometric algorithm in complex scenarios.

	\begin{table*}[tp]
		\centering
		\small
		\caption{Performance comparison of algorithms on Accuracy with IoU (\%) on the Cornell dataset.}
		\label{tab:performance}
		\begin{tabular}{p{4cm} p{6cm} p{3cm} p{1cm} p{0.5cm}}
			\toprule
			\textbf{Authors} & \textbf{Algorithm} & \textbf{Accuracy} & \multicolumn{2}{c}{\textbf{Labels}} \\
			\cmidrule(lr){4-5}
			\textbf{} & \textbf{} & \textbf{} & \textbf{Train} & \textbf{Test} \\
			\midrule
			\cite{b5}  & GG-CNN 2              & 84.0   & GT & GT \\
			\cite{b73} & ROI-GD, ResNet-101   & 93.6   & GT & GT \\
			\cite{b71} & DD-Net, Hourglass    & 97.2   & GT & GT \\
			\cite{b74} & SE-ResUNet           & 98.2   & GT & GT \\
			\cite{b64} & GKNet                & 96.9   & GT & GT \\
			\cite{b6}  & GR-ConvNet           & 98.8   & GT & GT \\
			\cite{b42} & HTC-Grasp            & 98.3   & GT & GT \\
			\cite{b91} & Grasp-DSC & 96.6 &  GT & GT \\
			\midrule
			This Work & SkelGG-CNN (Jacquard-trained) & 97.1   & StSkel & GT \\
			& SkelGG-CNN (Dex-Net-trained)  & 96.26  & StSkel & GT \\
			\bottomrule
		\end{tabular}
	\end{table*}

	\subsection{Ablation Study}
	Three ablation studies are performed on the model to demonstrate the effectiveness of the developed architecture.
	First, the model's performance is evaluated using different input types, including depth images, RGB images, and combined RGB-D images, to analyze their influence on the results.
	For the RGB-D input, the depth map is stacked as an additional channel alongside the RGB channels, forming a four-channel input.
	The reason for using RGB-D images is that object textures, especially along border lines, provide valuable information for the model to learn.
	Additionally, some transparent or reflective objects often produce unreliable depth maps.
	Table \ref{tab:rgbd_comp} shows accuracy results on Jacquard V1 dataset based on the input type of the model.
	The trained model with RGB-D input has the best performance, achieving 95.42\% accuracy with IoU metric on Jacquard V1 dataset.
	
	In a subsequent experiment, the number of trainable parameters in the model was reduced to investigate its data efficiency in learning from the generated grasps.
	Table \ref{tab:size_cmp} presents a comparison of models with varying parameter counts.
	Three models were trained, each with half the number of filters per layer compared to the previous configuration, resulting in the smallest model containing approximately one million parameters.
	Despite the reduced model size, the performance degradation was minimal, with the model achieving a grasp accuracy of 92.10\%.
	This resilience in performance can be attributed to the use of skeleton heatmaps, which effectively guide the model toward learning high-quality grasp regions. 
	Since the skeleton-based grasp keypoint algorithm generates grasps exclusively within these meaningful regions, the model concentrates on more informative areas, thereby enhancing learning efficiency.
	
	Finally, and most importantly, the effectiveness of the developed architecture must be evaluated in terms of both learning efficiency and grasp accuracy.
	For this purpose, the training and test images obtained from the gathered Dex-Net dataset were used. The SkelGG-CNN model was trained under two configurations: one without the skeleton output head (the conventional approach in existing literature) and another with the the skeleton head included.
	For comparison, the test results from the GG-CNN 2 model were also incorporated.

	Figure~\ref{fig:ablation_plot} illustrates the training performance across 100 epochs for the different architectures. The training and test datasets, generated using the proposed auto-labeling algorithm, indicate that the baseline architecture achieves better learning behavior than the original GG-CNN model. Furthermore, incorporating the skeleton heatmap leads to improved accuracy after several training iterations.
	These findings demonstrate that the geometric information captured in the skeleton heatmap enables the model to learn more effective grasp poses, ultimately resulting in superior performance.

	\begin{table*}[tp]
		\centering
		\footnotesize
		\caption{Detailed results on the GraspNet dataset (Kinect), showing average precision (AP) scores.}
		\label{tab:graspnet_compariosn}
		\begin{tabular}{lccccccccc}
			\toprule
			Methods & \multicolumn{3}{c}{Seen} & \multicolumn{3}{c}{Unseen} & \multicolumn{3}{c}{Novel} \\
			\cmidrule(lr){2-4} \cmidrule(lr){5-7} \cmidrule(lr){8-10}
			& AP (\%) & AP$_{0.8}$ (\%) & AP$_{0.4}$ (\%) & AP (\%) & AP$_{0.8}$ (\%) & AP$_{0.4}$ (\%) & AP (\%) & AP$_{0.8}$ (\%) & AP$_{0.4}$ (\%) \\
			\midrule
			GPD~\cite{GDP2017}         & 24.38 & 30.16 & 13.46 & 23.18 & 28.64 & 11.32 & 9.50 & 10.14 & 3.16 \\
			GG-CNN~\cite{b69}      & 16.89 & 22.47 & 11.23 & 15.05 & 19.76 & 6.19 & 7.38 & 8.78 & 1.32 \\
			PointNetGPD~\cite{Pointnetgpd} & 27.59 & 34.21 & 17.83 & 24.38 & 30.84 & 12.83 & 10.66 & 11.24 & 3.21 \\
			GraspNet~\cite{b63}   & 29.88 & 36.19 & 19.31 & 27.84 & 33.19 & 16.62 & 11.51 & 12.92 & 3.56 \\
			RGBMatter~\cite{RGBMatters}  & 32.08 & 39.46 & 20.85 & 30.40 & 37.87 & 18.72 & 13.08 & 13.79 & 6.01 \\
			TransGrasp~\cite{transgrasp} & 35.97 & 41.69 & 31.86 & 29.71 & 35.67 & 24.19 & 11.41 & 14.42 & 5.84 \\
			GraNet~\cite{GraNet}     & 41.48 & 49.84 & 33.86 & 35.29 & 43.15 & 26.89 & 11.57 & 14.31 & 5.24 \\
			HGGD~\cite{hggd}         & 61.17 & 69.82 & 56.52 & 47.02 & 56.78 & 38.86 & 19.37 & 23.95 & 12.14 \\
			EconomicGrasp~\cite{economic}   & 68.21 & 79.60 & 63.54 & 61.19 & 70.63 & 53.77 & 25.48 & 31.46 & 13.85 \\
			\midrule
			\textbf{Quasi-3D Auto-Labeling}         & 38.22 & 48.45 & 20.98 & 33.76 & 42.19 & 17.64 & 18.33 & 22.83 & 9.02 \\
			\bottomrule
		\end{tabular}
	\end{table*}

	\subsection{Evaluation on Jacquard V1 with Ground-Truth Testing}
	Grasp accuracy serves as one of the most important metrics for evaluating the grasping capabilities of robotic models.
	This metric provides a standardized approach to compare different methods, making it a popular benchmark in the field.
	However, while reporting accuracy on benchmarks such as the Jacquard dataset ensures a unified way to evaluate performance, it does not necessarily reflect the model’s robustness or real-world applicability.
	To evaluate the effectiveness of the proposed auto-labeling algorithm and the SkelGG-CNN model, the Jacquard dataset was processed and labeled using the straight skeleton grasp generation algorithm.
	The SkelGG-CNN model was trained on these auto-generated labels and assessed on the Jacquard V1 test set using its original ground-truth labels.
	
	This approach resulted in a grasp accuracy of 95.42\% on the Jacquard V1 test set, positioning SkelGG-CNN among the top-performing grasping models on this dataset, as shown in Table \ref{tab:jacq_comp}. 
	In comparison, state-of-the-art methods such as those in \cite{b74} and \cite{b71} achieve high accuracy by training directly on the ground-truth labels of the Jacquard dataset.
	In contrast, SkelGG-CNN’s competitive performance of 95.42\%, achieved using only algorithmically generated labels, highlights the robustness and quality of the proposed auto-labeling approach.
	Moreover, the auto-labeled grasps were compared with the ground-truth annotations, achieving an accuracy of 89.2\%, which indicates a high degree of similarity to the original labels.
	
	As previously discussed, one of the motivations for employing a neural network rather than relying solely on the auto-labeling approach is that the labeling algorithm does not provide any estimation of grasp quality.
	In contrast, SkelGG-CNN not only predicts the most confident grasp poses but also learns grasp patterns across geometries frequently encountered during training.
	This enables the model to effectively filter out noisy or incorrect grasps that may have been introduced during the auto-labeling process for some geometries, resulting in cleaner and more reliable grasp predictions.
	
	Figure \ref{fig:jacq_fig_comp} illustrates the model's grasp predictions on various images from the Jacquard V1 test set, with grasp configurations represented as rectangles.
	The model outputs are compared with those of the state-of-the-art Gr-ConvNetV2 \cite{b6} method, which clearly illustrates the impact of the skeleton heatmap on the predicted grasps.
	These results demonstrate the model's ability to generate diverse and accurate grasp poses across different object regions.
	The high accuracy achieved by training on auto-generated labels and testing against ground-truth labels underscores the strength and validity of the auto-labeling algorithm.
	It demonstrates that the algorithm produces high-quality annotations capable of training a model that performs comparably to those trained on manually labeled data.
	
	This result is particularly significant as it confirms the framework’s ability to auto-label data, train a robust model, and achieve strong generalization, as validated by established benchmark standards.
	The comparison between the proposed method and state-of-the-art approaches is conducted solely for demonstration purposes. Models trained on ground-truth labels have an advantage, as they tend to predict grasps similar to those encountered during the training phase.
	
	Experimental results demonstrate that the developed model effectively predicts a diverse range of grasp poses across various regions of the objects.
	Moreover, integrating a grasp center correction mechanism using a skeleton heatmap significantly enhances grasp quality by aligning the grasp centers with locally optimal centers and mitigating errors in grasp width estimation.
	The issue of incorrect width or orientation estimation in generative grasp models arises because the model tends to predict the orientation and width of a grasp rectangle with higher confidence near the edges of the object.
	As a result, prediction accuracy significantly decreases across all heatmaps toward the edges of the grasp rectangles, leading to less precise grasp estimations.
	This issue becomes more pronounced when the grasp rectangle is larger, as the predictions tend to produce uneven and disjointed regions across the grasping rectangle, leading to inconsistencies in grasp estimation.
	
	The failure cases of the model, as measured by the IoU metric, were primarily due to insufficient grasp coverage for certain objects in the Jacquard dataset, a limitation also noted by \cite{b64}. 
	The remaining failure cases were primarily attributed to the limitations of the grasp annotation algorithm.
	Specifically, the grasp poses generated by the straight skeleton method consider only the front-facing polygon of objects, neglecting variations in elevation and 3D geometry.
	Despite these limitations, both the grasp keypoint algorithm and SkelGG-CNN have been successfully deployed for precise and reliable grasping tasks, demonstrating their effectiveness in practical applications.
	
	As an additional experiment, the SkelGG-CNN model trained on Dex-Net auto-labeled data was tested on the Jacquard V1 ground-truth labels, achieving a grasp accuracy of 94\%. This result, while not part of the core Jacquard evaluation strategy, underscores the model’s ability to generalize across datasets.

	\subsection{Evaluation on Jacquard V2 with Ground-Truth Testing}\label{subsec:eval_jacquardV2}
	
	The Jacquard V2 dataset, as detailed in \cite{b75}, builds upon the original Jacquard V1 dataset by introducing a Human-In-The-Loop (HIL) approach to refine its annotations.
	Unlike Jacquard V1, which generates all grasp labels automatically and often suffers from problems such as missing or incorrect grasp bounding boxes, limited grasp orientations, and incomplete annotations, Jacquard V2 includes human operators who selectively correct the automated labels. This refinement process produces labels that are significantly more akin to those created by human experts, establishing Jacquard V2 as a more accurate benchmark for evaluating grasp detection performance.
	
	In this evaluation, the SkelGG-CNN model was trained exclusively on grasp labels generated automatically by the straight skeleton and pairing method, without any human intervention. The model was then tested on the human-refined Jacquard V2 dataset, achieving an improvement in grasp accuracy from 95.42\% (on Jacquard V1) to 96.55\%. This increase in performance is noteworthy because it demonstrates that the auto-generated labels, used during training, are highly similar to the human-refined annotations in Jacquard V2. The ability of the SkelGG-CNN model to achieve such high accuracy on a dataset with human-level ground-truth labels underscores the quality of the proposed auto-labeling algorithm. The primary purpose of this evaluation is to assess how closely the auto-generated labels approximate human-level annotations. The strong performance on Jacquard V2, where the model achieves 96.55\% accuracy and outperforms several existing methods as shown in Table~\ref{tab:jacquard_V2}, validates that the automatic labeling process produces annotations nearly equivalent to those refined by humans.

	\subsection{Cross-Dataset Generalization Evaluation on Cornell}
	To evaluate the generalization capabilities of the SkelGG-CNN model across diverse datasets, two instances of the model were trained using auto-labeled data: one on the Jacquard dataset and the other on the Dex-Net dataset. Both models were subsequently tested on the Cornell dataset, which comprises 885 manually annotated real-world RGB-D images of 240 distinct objects.
	The human-labeled ground-truth annotations in the Cornell dataset provide a benchmark for assessing the model's ability to perform effectively on data distributions different from those encountered during training.
	
	The SkelGG-CNN model trained on the Jacquard dataset achieved a grasp accuracy of 97.1\% on the Cornell dataset, as measured by the Intersection over Union (IoU) metric with a threshold of 0.25 and an orientation error less than $\pi/6$. In parallel, the model trained on the Dex-Net dataset attained a grasp accuracy of 96.26\%.
	These high accuracies achieved on a dataset with human-annotated labels demonstrate that the automatic labeling process generates annotations closely approximating human-level quality, reducing reliance on extensive manual labeling.
	The near-equivalent performance of the Dex-Net-trained model, despite its significantly smaller training dataset, is particularly striking, with only a marginal accuracy difference of 0.84\% compared to the Jacquard-trained model.
	This cross-dataset evaluation serves two primary purposes:

	\textbf{Testing Generalization Across Datasets:} The experiment assesses the SkelGG-CNN framework’s ability to generalize across varying data distributions. By training on auto-labeled data from one dataset (e.g., Jacquard or Dex-Net) and evaluating on the Cornell dataset with its human-annotated labels, the results validate the robustness of both the auto-labeling algorithm and the SkelGG-CNN model. The accuracies of 97.1\% (Jacquard-trained) and 96.26\% (Dex-Net-trained) highlight the framework’s adaptability to real-world scenarios without necessitating dataset-specific retraining or manual labeling. This capability is critical for deploying robotic grasping systems in diverse, unstructured environments where pre-labeled data may not be available.
	
	\textbf{Demonstrating Efficiency with Small Datasets:} The comparable performance of the Dex-Net-trained model underscores the efficiency of the geometry-aware learning approach embedded in SkelGG-CNN. Despite the Dex-Net dataset being over 20 times smaller than Jacquard (2,622 vs. 54,000 images), the model achieves nearly equivalent accuracy. This efficiency arises from the integration of geometric features via the skeleton heatmap, which guides the model to focus on high-quality grasp regions and learn effective grasp strategies with fewer training samples. The ability to maintain robust and generalized performance with limited data enhances the framework’s scalability and practicality for new tasks or environments where data collection is constrained.
	
	The performance of SkelGG-CNN is compared with state-of-the-art models in Table~\ref{tab:performance}. 
	The Jacquard-trained model’s 97.1\% accuracy and the Dex-Net-trained model’s 96.26\% accuracy position SkelGG-CNN competitively among recent methods, many of which leverage advanced architectures like Transformers (e.g., GraspFormer, TransGrasp) or deeper networks (e.g., SE-ResUNet).
	However, unlike these works, which often prioritize architectural scalability, this study emphasizes the effectiveness of the proposed grasp generation approach using algorithmically generated labels.
	The high accuracies achieved without human-labeled training data affirm the model’s learning efficiency.
	
	These results confirm that the SkelGG-CNN framework can efficiently auto-label data, train a robust model, and generalize effectively to unseen datasets.
		\begin{figure}[tp]
		\centering
		\captionsetup{}
		\includegraphics[width=\linewidth]{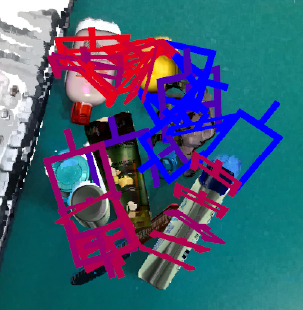}
		\caption{Results of the quasi-3D auto-labeling method on a real-world scene from the GraspNet dataset, visualized with a blue-to-red gradient indicating grasp quality (red: higher quality, blue: lower quality).}
		\label{fig:graspnet_exp}
	\end{figure}

	\begin{figure*}[tp]
		\centering
		\captionsetup{}
		\includegraphics[width=\linewidth]{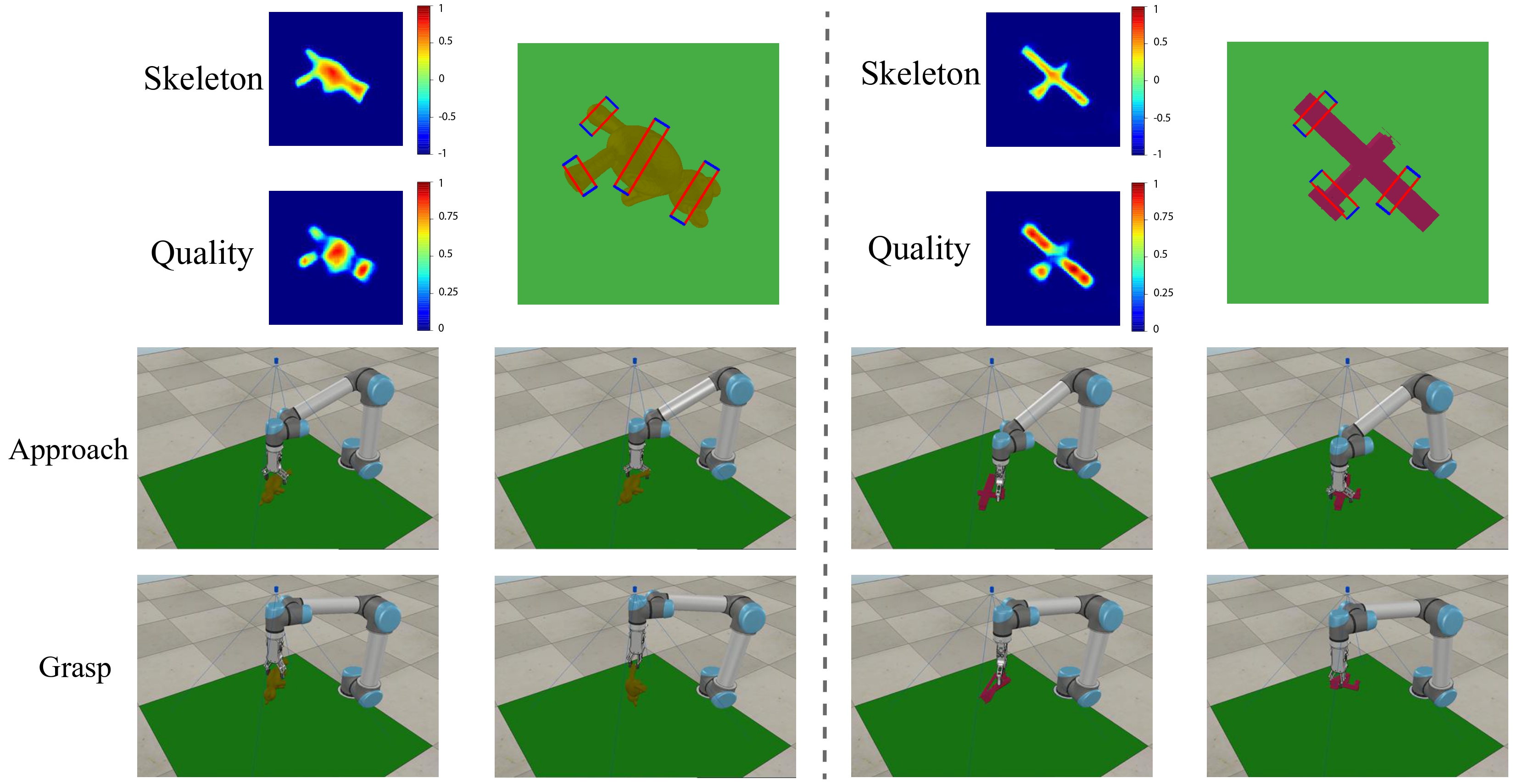}
		\caption{Simulation experiments on the Dex-Net dataset. 
			The first row shows the predicted grasp rectangles generated by the SkelGG-CNN model, along with the corresponding skeleton and quality heatmaps.
			The second row illustrates the robotic arm's approach phase.
			Finally, the third row presents the successfully grasped objects from two different grasps.}
		\label{fig:sim_results}
	\end{figure*}

	\subsection{Generalized 6-DoF Grasp Annotation on GraspNet benchmark}\label{subsec:graspnet}
	The performance of the generalized quasi-3D auto-labeling algorithm, introduced in Section~\ref{subsec:6dof_method}, is evaluated on the GraspNet benchmark. GraspNet uses a force-closure metric with different friction coefficients to assess grasp quality, making it a suitable testbed for evaluating 6-DoF grasp generation methods on complex, cluttered point cloud scenes.
	
	Table~\ref{tab:graspnet_compariosn} presents the detailed results of the proposed quasi-3D auto-labeling algorithm evaluated on the GraspNet dataset (Kinect), reporting the average precision (AP) scores. 
	Evaluations were conducted solely on the test dataset, which includes seen, unseen, and novel object categories.
	All compared methods are model-based approaches trained on GraspNet annotations, with performance largely driven by learned scoring networks that estimate grasp quality.
	These scoring models play a critical role in achieving high AP scores.

	As shown in Table~\ref{tab:graspnet_compariosn}, the proposed method achieves AP scores of 38.22, 33.76, and 18.33 for seen, unseen, and novel categories, respectively. 
	In the seen and unseen categories, which test intra-category generalization (grasping unseen instances of known categories), the AP scores are competitive but do not rank among the top three performers, which can be attributed to the absence of a learned scoring module.
	In contrast, in the novel category, which evaluates cross-category generalization (grasping entirely unseen object types) and presents greater difficulty, the method achieves an AP of 18.33, showing high scores among compared methods.
	
	It is acknowledged that the current approach does not leverage a dedicated scoring model, which limits the achievable precision compared to model-based counterparts. 
    As a direction for future work, the framework can be extended by integrating a more sophisticated, model-based scoring module that incorporates object skeleton features.
    Subsequently, a valuable direction for future work would be to fully integrate the extracted 3D point-cloud features into a grasp-generation model, although this remains challenging due to the incomplete and unstructured nature of point-cloud data.
	Such an enhancement is expected to improve grasp quality evaluation and overall performance.
	
	The learned component of the framework, SkelGG-CNN, is not evaluated on the GraspNet benchmark, as it is designed for 4-DoF grasp generation, whereas GraspNet is specifically tailored for evaluating 6-DoF grasp configurations.
	Consequently, models in the GG-CNN family are not typically assessed under this benchmark due to this fundamental mismatch. 
	The evaluation in this section instead focuses on providing a fair assessment of the proposed generalized skeletonization algorithm, demonstrating its capability to achieve competitive performance compared to state-of-the-art 6-DoF methods. 
	An example of generated grasp labels using the quasi-3D auto-labeling algorithm is shown in Figure~\ref{fig:graspnet_exp}.

	\subsection{End-to-End Framework Evaluation on Dex-Net (Simulation Experiments)}\label{sec:subsec54}
	
	The data collection and grasping experiments were conducted using the CoppeliaSim robotic simulation tool, providing a high-fidelity environment for robotic manipulation research. 
	A UR5 6-DoF robot equipped with an RG2 gripper was employed to simulate grasping tasks across a range of object geometries. 
	The test and unseen subsets of Dex-Net objects were placed in the simulation workspace. 
	For each predicted grasp configuration, the robot executed the grasp, and a trial was considered successful if the object was lifted and held steadily for three seconds.

	The SkelGG-CNN model achieved outstanding grasp success rates of 97.6$\%$ on the test set and 95.2$\%$ on unseen objects.
	These results validate the model’s ability to generalize to both familiar and novel shapes, despite being trained on a relatively limited dataset. 
	Figure~\ref{fig:sim_results} illustrates several representative examples of successful grasp predictions within the simulated environment. 
	For further evaluations, the grasp success rates of two 4-DoF state-of-the-art models along with the auto-labeling results are measured and compared.
	The 2D auto-labeling approach achieves a 92.8\% success rate on the test set and 88.0\% on the unseen object set, which is comparable to model-based approaches.
	
	Furthermore, the quasi-3D auto-labeling approach outperforms all these methods in the case of grasp success rate, with 98.4 on the test set and 96.0 on the unseen object set.
	The generalized auto-labeling shows superior results in this benchmark than all generative models and the 2D auto-labeling approach.
	The failure cases indicate that the SkelGG-CNN model mainly suffers from depth estimation in many cases.
	Because the model does not use the data from the point cloud, the depth estimation method, which uses the depth of the center of the grasp in some cases results in the robot trying to go through the object.
	The other problem is that top-view images do not fully capture the 3D volume of the object, which results in the model not having sufficient data for the grasp.
	These two reasons form the main failure cases of the SkelGG-CNN and the 2D auto-labeling approach, which are mostly avoided in the case of the 6-DoF grasp generation algorithm.
	
	The evaluation of the 6-DoF method demonstrates an improved success rate across all test sets, indicating that this approach can generate high-quality grasps even for complex geometries.
	A potential concern regarding the developed algorithm is that the approximations of 2D polygons derived from point cloud cross-sections using the alpha-shape algorithm might introduce inaccuracies. 
	However, the evaluation results confirm that this is not the case. 
	Even in scenarios where 2D skeletonization fails to fully capture the cross-sectional areas, the centers of the skeleton nodes still accurately correspond to the true grasp region centers.

\begin{table}[tp]
	\centering
	\scriptsize
	\setlength{\tabcolsep}{4pt}
	\caption{Success rate comparison of algorithms on Dex-Net and EGAD! datasets.}
	\label{tab:dex_egad}
	\begin{tabular}{llccc}
		\toprule
		\textbf{Category} & \textbf{Method} & \textbf{EGAD!} & \textbf{Dex-Net Test} & \textbf{Dex-Net Unseen} \\ 
		\midrule
		\multirow{3}{*}{Learning} 
		& GG-CNN          & 54.0 & 89.6 & 88.0 \\
		& GR-ConvNetV2     & -    & 94.4 & 92.2 \\
		& SkelGG-CNN$^{*}$ & 62.4 & 97.6 & 95.2 \\     
		\midrule    
		\multirow{2}{*}{Geometric} 
		& 2D Auto-Labeling & 60.8 & 92.8 & 88.0 \\
		& Quasi-3D Auto-Labeling & 89.1 & 98.4 & 96.0 \\
		\bottomrule
	\end{tabular}
	
	\vspace{2pt}
	{\footnotesize $^{*}$SkelGG-CNN is trained using grasp labels generated by the 2D Auto-Labeling method.}
\end{table}

	\subsection{Evaluation on EGAD! Dataset for Measuring Limitations of 4-DoF Grasping Methods}
	In this setup, objects from the evaluation set of EGAD! are placed in the same configuration as described in Section~\ref{sec:subsec54}. 
	As noted previously, the challenges of depth estimation and the lack of complete geometric information make 4-DoF grasping approaches particularly prone to failure. 
	The EGAD! dataset is specifically designed to expose such limitations, as the complex object geometries often lead to unstable grasps when only partial 3D information is available.
	For this reason, the EGAD! dataset serves as a valuable benchmark for evaluating the robustness of both 4-DoF and 6-DoF grasp generation methods. 
    For the SkelGG-CNN model, which is trained using grasps generated by the 2D auto-labeling method, the top five predicted grasp configurations are selected for evaluation. 
    If the model produces fewer than five grasp poses with a confidence score below 0.8, the experiment is repeated up to five times. For the 2D auto-labeling approach, which serves as the teacher for SkelGG-CNN, five random grasp configurations are chosen for each object, whereas for the quasi-3D auto-labeling approach, the top five grasps with the highest scores are selected.

	Table \ref{tab:dex_egad} presents the success rates of several learning-based and geometry-based approaches. 
    As shown, the proposed SkelGG-CNN achieves strong performance on the Dex-Net benchmarks, obtaining success rates of 97.6\% on the Dex-Net Test set and 95.2\% on the Dex-Net Unseen set. 
    These results outperform the baseline GG-CNN and are competitive with more complex architectures such as GR-ConvNetV2.
    The results indicate that the geometry-aware supervision generated by the proposed 2D auto-labeling pipeline enables the model to learn robust grasp representations. 
    The 2D auto-labeling geometric method itself achieves a 60.8\% success rate on the EGAD! benchmark.
    When trained on these automatically generated labels, SkelGG-CNN achieves a slightly higher success rate of 62.4\%, demonstrating that the learned model is able to match and marginally improve upon the performance of its 2D teacher algorithm while providing the advantages of fast inference and better generalization in practical robotic scenarios.

	\begin{figure}[tp]
		\centering
		\captionsetup{}
		\includegraphics[width=\linewidth]{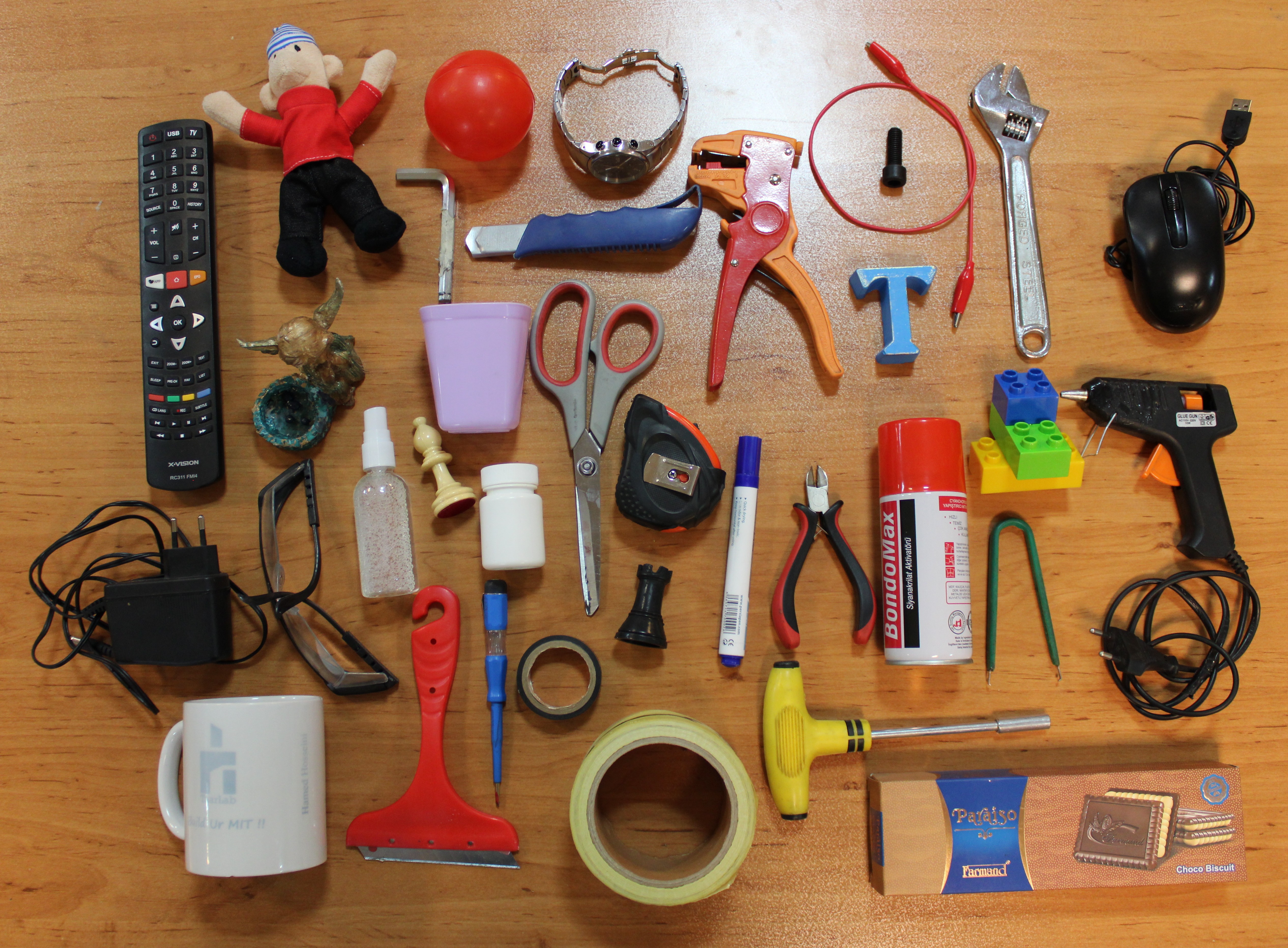}
		\caption{Illustration of the objects used for real robotic experiments.}
		\label{fig:objects}
	\end{figure}

	While 4-DoF methods perform well on objects with relatively uniform geometry along their height, they struggle in scenarios where volumetric variations require reorienting the gripper to maintain stable contact. In contrast, 6-DoF approaches can adjust both position and orientation in 3D space, enabling adaptation to irregular geometries and consistently achieving robust and stable grasps. The results demonstrate that the generalized 3D grasp generation algorithm significantly outperforms both SkelGG-CNN and its 2D auto-labeling teacher, achieving a success rate of 89.1\%, demonstrating the potential of the proposed skeleton-based method when full 3D geometric information is available.
	
		\begin{figure*}[tp]
		\centering
		\captionsetup{}
		\includegraphics[width=0.75\linewidth]{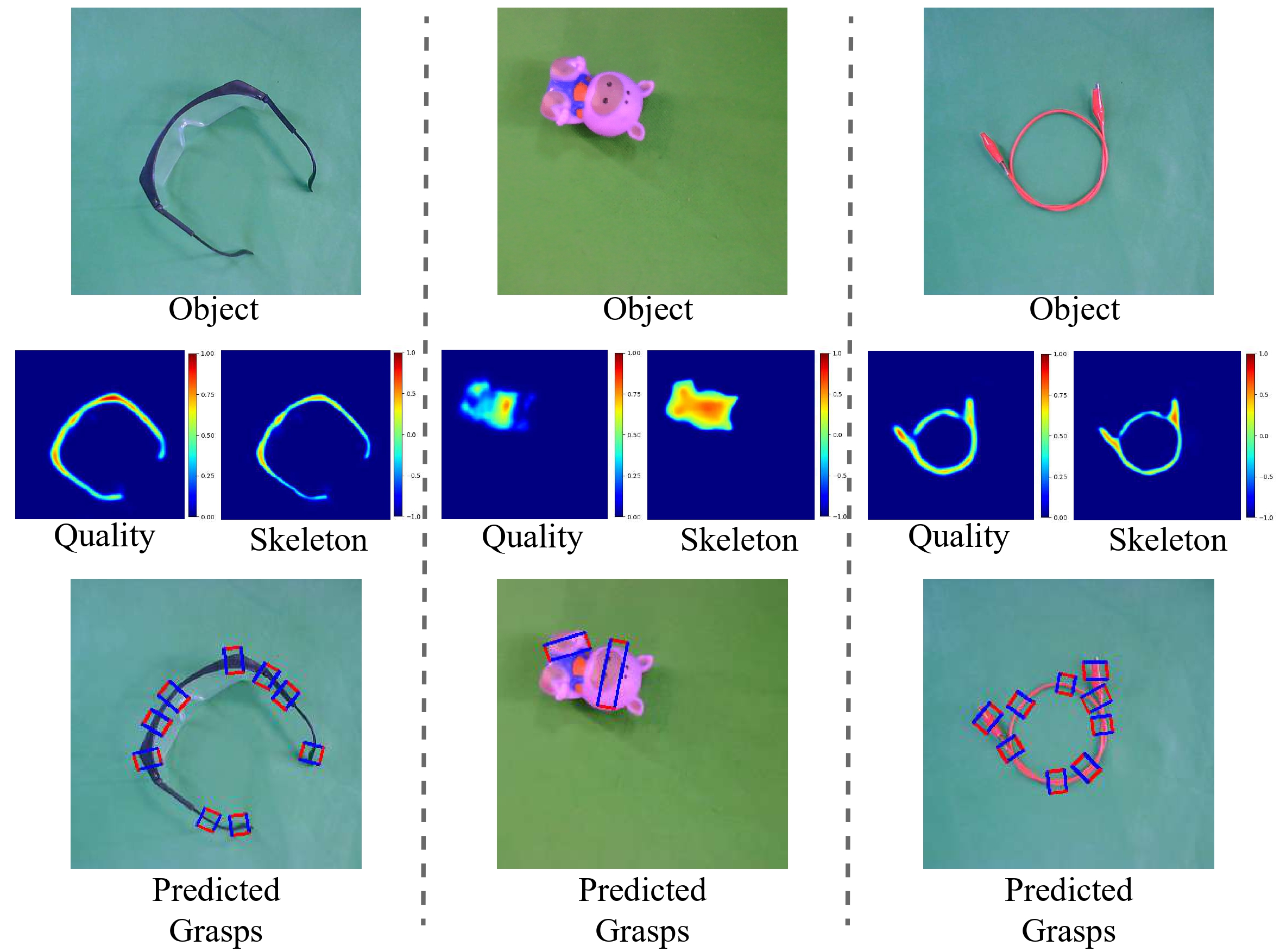}
		\caption{Grasp predictions produced by the SkelGG-CNN model for single objects in a real-world environment. The figure displays both quality and skeleton maps, highlighting the identified grasp configurations.}
		\label{fig:single1}
	\end{figure*}

	\section{Physical Experiments}
	\label{sec:section6}
	Evaluating a grasping model solely against benchmarks is insufficient to fully validate its robustness, as real-world conditions introduce challenges such as sensor noise, occlusions, and dynamic object interactions. To comprehensively assess the capabilities of the SkelGG-CNN model, a robotic setup was implemented, and two sets of experiments were conducted. These experiments aimed to demonstrate the model's effectiveness in practical grasping tasks, validating its performance beyond controlled benchmark datasets, and highlighting its adaptability to noisy and cluttered real-world environments.
	
	In this evaluation, the SkelGG-CNN model, trained exclusively on the Dex-Net dataset in simulation using single objects, was tested. The objective is to investigate the model's generalization capabilities to real-world environments, particularly in cluttered scenes, without employing sim-to-real transforms or domain adaptation methods. This assessment underscores the model’s ability to extend its performance from simulated single-object training to complex, unstructured real-world scenarios, a critical factor for its practical deployment.
	
	The available hardware platform at the Human and Robot Interaction Laboratory is a 3-DoF Delta parallel robot equipped with a 1-DoF gripper, resulting in a total of 4 controllable degrees of freedom.
	While generalized grasping algorithms are often formulated for 6-DoF, this level of control is not strictly necessary for many industrial pick-and-place tasks.
	In structured environments such as factory lines, where objects are typically presented on conveyors or in bins with predictable orientations, planar positioning combined with gripper actuation is sufficient to achieve reliable manipulation.
	The Delta robot, in particular, offers very high precision and repeatability within its workspace, making it well-suited to these industrially relevant scenarios.
	
	The focus of this work is the development of a 2D auto-labeling algorithm that operates directly on RGB-D images.
	To highlight this contribution, physical experiments are restricted to 4-DoF pick-and-place tasks on the Delta robot, which faithfully represent many real-world automation settings.
	Broader evaluations of 6-DoF grasping are conducted in simulation, serving as a proof of generalization beyond the limitations of the available hardware.
	
	\subsection{Robotic Environment Setup}
	The experimental setup, as shown in Figure \ref{delta}, employs a 3-DoF Delta parallel robot~\cite{b84} equipped with a two-fingered gripper~\cite{b83} mounted on its end-effector. 
	A PID control scheme ensures precise positioning of the Delta robot.
	The operational sequence initiates from a defined home configuration, followed by movement to a location directly above the target object.
	Alignment is achieved based on the orientation specified by the grasp detection algorithm, after which the robot attempts to grasp the object from the specified configuration.
	The grasping process is refined by a control algorithm that closes the gripper as it approaches the object until an appropriate force is detected by an FSR sensor mounted on the gripper plane. Additionally, two cameras are utilized in the experimental setup: an RGB camera for recording color images and a Kinect V1 sensor for capturing depth images.
	The calibration procedure for the Kinect sensor is carried out using the same method adopted in~\cite{b15}.
	
	\begin{figure*}[tp]
		\centering
		\captionsetup{}
		\includegraphics[width=\linewidth]{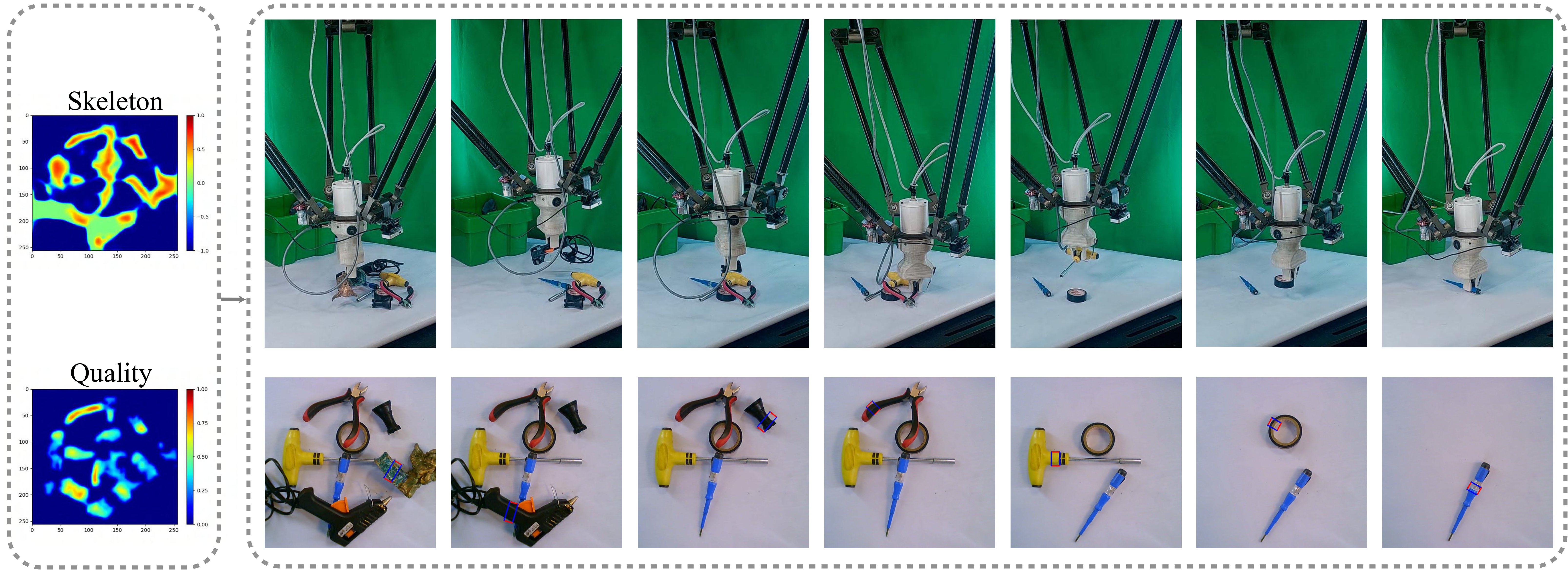}
		\caption{Demonstration of a Delta parallel robot grasping objects in a cluttered scene. The first column presents the skeleton and quality output of the SkelGG-CNN model for the initial scene. For each grasp attempt, grasps are generated based on the captured image of the scene in its current state. The second row highlights the bounding box of the selected grasp at each stage.}
		\label{fig:cluttered1}
	\end{figure*}

	\subsection{Grasping Single Objects}
	In the first experiment, a set of 35 household objects with varying shapes, sizes, and materials was selected to evaluate the model’s performance in a diverse grasping scenario.
	The objects included both rigid and soft items, some resembling those in the training dataset (e.g., cups and boxes), while others, such as gloves, were novel to the model.
	During the experiment, the camera was fixed above the scene, and objects were randomly placed within the robot’s workspace at various positions and orientations.
	After grasp pose detection, the predicted grasp configurations were transformed into the robot’s reference frame for execution.
	The robot then attempted to grasp each object and hold it without slipping for three seconds.
	A grasp was considered successful if the object remained securely held, while object slipping or dropping was considered a failure.
	Each object was tested 10 times, resulting in 350 attempts.
	Figure \ref{fig:objects} illustrates the selected objects used in this experiment.
	Despite being trained solely on simulation data, the SkelGG-CNN model achieved a 97.4\% grasp success rate, demonstrating reliable performance in grasping tasks.
	The high success rate validates the effectiveness of the geometry-aware framework, enabling generalization to real-world conditions despite challenges like sensor noise and varying object properties. Figure \ref{fig:single1} illustrates the grasp predictions of the SkelGG-CNN model for single objects in a real-world setting, displaying both the quality map and the skeleton map.
	Failure cases were primarily linked to objects lacking uniform geometry along the z-axis. 
	In such instances, even when the model accurately predicted the grasp pose, the robot could not secure a stable grip from a top-down approach, resulting in a failed grasp.
	Additionally, in rare cases, the predicted grasp points were positioned too close to the object’s endpoint, reducing the contact area between the gripper and the object, which occasionally caused slipping.
	
	\subsection{Grasping in Cluttered Scenes}
	
	A key strength of the SkelGG-CNN model, which utilizes a generative grasping approach with convolutional layers, lies in its ability to generalize from single-object scenarios to cluttered environments, even when trained exclusively on isolated objects.
	Convolutional layers enable the model to identify graspable regions based on local and global geometric features, enabling the architecture to be capable of generating accurate pixel values in cluttered scenes.
	To test this capability, ten test scenes were created, each containing a mix of randomly arranged objects.
	Objects were randomly placed by shaking them in a box and emptying them into the robot’s workspace, producing diverse, unstructured layouts for each scene.
	In this experiment, the robot attempted to grasp each object and place it into a designated box at a fixed location within the workspace. 
	A scene was considered successfully completed when all objects from the initial arrangement had been picked up and placed into the box.
	Performance was evaluated using both grasp success rate (successful pick-ups) and task completion rate.
	The task completion rate was computed as the average percentage of objects successfully grasped and placed per scene, where each scene’s percentage is the number of successfully handled objects divided by the total objects in that scene.
	The cluttered experiments involved ten test scenes, each with a varying number of objects.
	For each scene, a top-view image was captured, and grasp selection was performed using the combination of quality score and skeleton heatmap to reduce the risk of collision during the grasp. 
	This process was repeated until all objects were cleared from the scene.
	Across the ten scenes, the SkelGG-CNN model achieved a grasp success rate of 95.0\% and a task completion rate of 97.0\%, showcasing its capability to handle cluttered scenes effectively despite being trained only on single objects. Figure~\ref{fig:cluttered1} demonstrates the robot grasping objects in a cluttered scene, with the model’s skeleton and quality outputs guiding the process.
	
\begin{figure}[tp]
		\centering
		\captionsetup{}
		\includegraphics[width=0.8\linewidth]{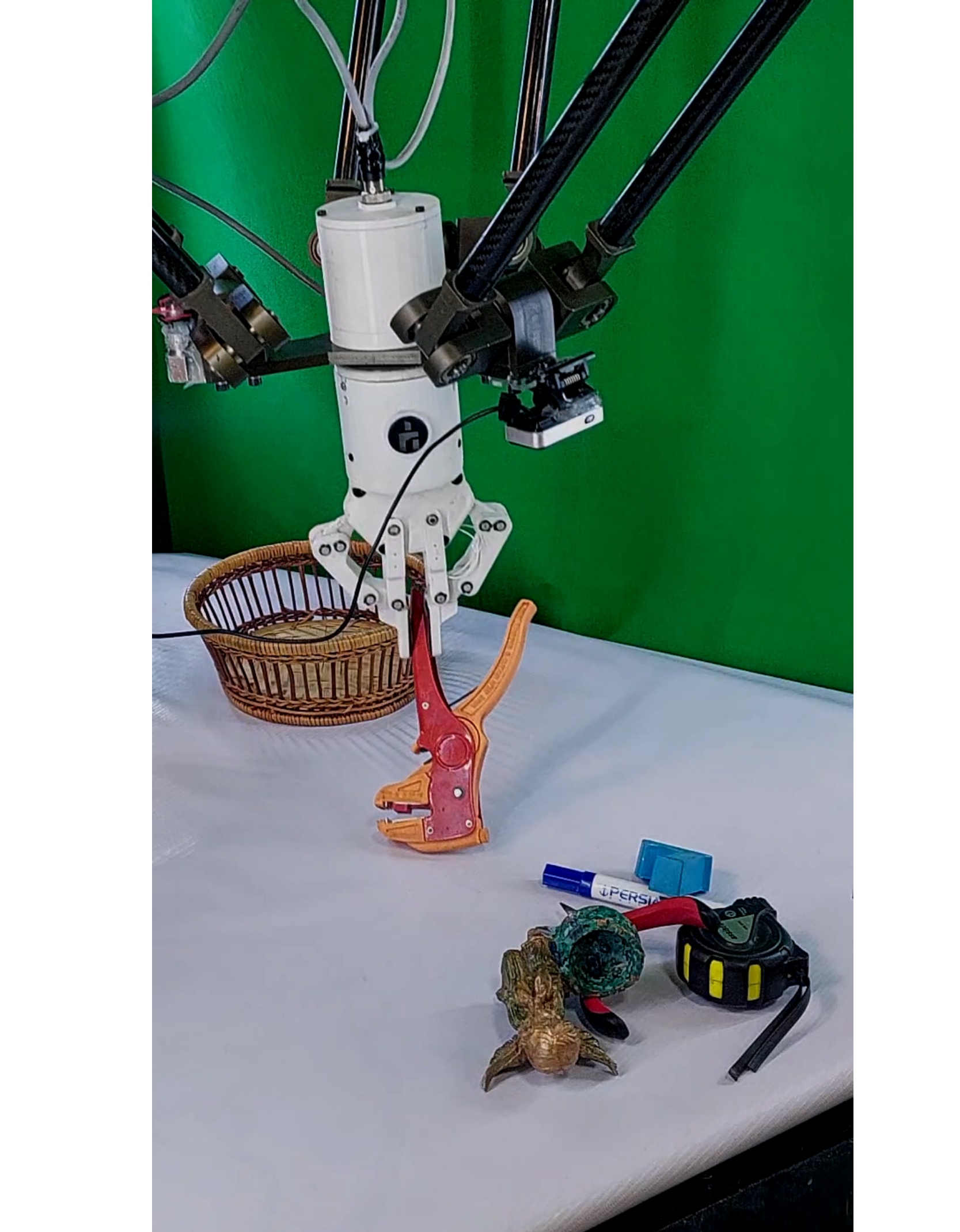}
		\caption{A failure case where the object slipped from the gripper after being grasped.}
		\label{fig:fail}
	\end{figure}

	A performance comparison with state-of-the-art models is presented in Table \ref{tab:household_performance}. Failure cases, such as the one shown in Figure~\ref{fig:fail}, occurred when the robot could not secure a firm grip, causing the object to slip from the gripper jaws despite a correctly predicted grasp pose. The results in Table \ref{tab:household_performance} indicate that the SkelGG-CNN model outperforms existing methods in both single-object and cluttered scene grasping tasks, achieving these outcomes without real-world training data or domain adaptation techniques. This success underscores the robustness of the model’s geometry-aware framework and its ability as a fully automatic pipeline to be used in different grasping scenarios.
	
	The high success rates in both isolated and cluttered scenes demonstrate that the SkelGG-CNN framework, from auto-labeling with straight skeletons to model training, effectively automates data preparation in simulation and delivers a model capable of tackling the complexities of real-world grasping tasks. These outcomes highlight the model’s adaptability to diverse object types and unstructured environments, reinforcing its potential for practical deployment.

	\begin{table*}[tp]
		\centering
		\small
		%\scriptsize
		\caption{Performance comparison on household isolated and cluttered scenes.}
		\label{tab:household_performance}
		\begin{tabular}{p{5.8cm} p{3.3cm} p{3.0cm} p{3.5cm}}
			\toprule
			\textbf{Algorithm} & \textbf{Training Dataset} & \textbf{Household Isolated} & \textbf{Household Cluttered} \\
			\midrule
			SAE, struct. reg. \cite{b14} & Cornell             & 89.0 (89/100)  & - \\
			AlexNet-based CNN \cite{b13} & Custom              & 66.0 (99/150)  & 38.4 (50/130) \\
			Robust Best Grasp \cite{b87} & ModelNet in Sim     & 80.0 (80/100)  & - \\
			Multi-grasp Res-50 \cite{b88} & Cornell             & 89.0 (89/100)  & - \\
			DexNet 2.0 \cite{b3}       & DexNet in Sim       & 80.0 (40/50)   & - \\
			GPD \cite{b60}              & CAD models          & -              & 77.3 (116/138) \\
			CTR \cite{b90}              & Custom in OpenRAVE  & 97.5 (39/40)   & 88.8 (66/74) \\
			GG-CNN \cite{b69}             & Cornell             & 91.6 (110/120) & 86.4 (83/96) \\
			GR-ConvNetV2 \cite{b6}        & Cornell             & 95.4 (334/350) & 93.5 (187/200) \\
			CFFGN \cite{b92}             & -             & 95.6 (239/250) & - \\\midrule
			This Work                             & DexNet in Sim & 97.4 (341/350) &  95.0 (135/142)\\ 
			\bottomrule
		\end{tabular}
	\end{table*}

	\section{Conclusion}
	\label{sec:conclusion}
	This research presented a novel framework for robotic grasp detection for two-finger grippers that integrated a geometry-aware grasp pose estimation algorithm, an automated data pipeline, and the Skeleton-based Generative Grasp Convolutional Neural Network (SkelGG-CNN) to address the challenges of grasping in diverse and unstructured environments. The framework leveraged straight skeletons to generate accurate grasp keypoints, thereby eliminating the need for manual labeling and significantly reducing computational complexity. Additionally, the auto-labeling method was extended to 6-DoF grasping through a generalized skeletonization approach, enabling volumetric grasp annotation directly on point clouds for more comprehensive handling of cluttered 3D scenarios. By incorporating geometric features through a skeleton heatmap, the SkelGG-CNN model achieved robust generalization and high grasp success rates across multiple benchmarks and real-world scenarios, thereby demonstrating its efficacy as a scalable solution for robotic manipulation.
	
	The experimental results affirmed the framework's superior performance. On the Jacquard V1 and V2 datasets, the model achieved grasp accuracies of 95.42\% and 96.55\%, respectively, using auto-generated labels, which highlighted the precision of the straight skeleton-based annotation method. Cross-dataset evaluations on the Cornell dataset further validated its generalization, with accuracies of 97.1\% and 96.26\% when trained on Jacquard and Dex-Net data, respectively. On the GraspNet benchmark, the 6-DoF auto-labeling achieved average precisions of 38.22 (seen), 33.76, and 18.33 (novel), demonstrating the strong generalization capabilities of the skeletonization method to 6-DoF auto-labeling.
	In simulation, the model attained a 97.6\% success rate on the Dex-Net test set and 95.2\% on unseen objects. The 6-DoF labeling further improved these rates to 98.4\% and 96.0\%, respectively, outperforming 4-DoF methods. On the EGAD! dataset, the 6-DoF approach yielded an 89.1\% success rate, addressing limitations of 4-DoF grasping on complex geometries. Physical experiments with a Delta parallel robot reinforced these findings and yielded a 97.4\% success rate for single-object grasping and a 95.0\% success rate in cluttered scenes, outperforming state-of-the-art methods without requiring real-world training data or domain adaptation.
	
	A key strength of the framework was its efficiency, as it achieved competitive performance with a relatively small dataset (2,622 images from Dex-Net) and a lightweight model architecture (approximately 15 million parameters in its largest configuration). This efficiency stemmed from the geometry-aware learning approach, in which the skeleton heatmap guided the model to focus on high-quality grasp regions, thereby enhancing learning with minimal data. The fully automated pipeline—from data collection and labeling to model training and deployment—represented a significant advancement toward practical, adaptable robotic grasping systems.
	
	In conclusion, the developed pipeline and use of skeletons in grasping offer a robust, efficient, and generalizable solution for robotic grasp detection. Its extension to 6-DoF grasping enhances its applicability to cluttered and complex 3D scenarios, bridging the gap between 4-DoF efficiency and full volumetric manipulation. Its innovative use of straight skeletons and geometry-aware learning paves the way for scalable robotic manipulation systems capable of operating effectively in various real-world settings.
	
	Although the current SkelGG-CNN architecture focuses on 4-DoF planar grasp prediction, the proven efficacy of the 6-DoF skeletonization algorithm on complex benchmarks like GraspNet establishes it as a highly reliable, automated annotation method. This bridges the gap between analytical and data-driven methods, providing the necessary foundation for training native 6-DoF point-cloud networks in future studies. However, a limitation remains in handling objects with complex 3D geometries, where some elevation variations and volumetric patterns can be neglected by the current 2D slicing heuristic.
    While this generalized quasi-3D algorithm was designed to ensure robustness and computational efficiency on partial point clouds, it may overlook some grasp-relevant features in complex spatial configurations. 
    Therefore, future research will focus on developing geometry-aware skeletonization algorithms that explicitly capture 3D topological structures.
    Additionally, future work could extend the overarching framework to accommodate multi-fingered grasps and integrate supplementary sensory inputs, such as tactile feedback, to further improve grasp stability and align more closely with physical grasping constraints.
	Exploring synergies with other geometric representations, such as medial axis transforms, may also refine grasp pose estimation. Finally, testing the framework's adaptability across diverse robotic platforms and environments will broaden its real-world applicability.

	\section*{Acknowledgments}
	The authors acknowledge that no additional contributors or third-party assistance were involved in this research.
	
	\section*{Declaration of conflicting interests}
	The authors declared no potential conflicts of interest with respect to the research, authorship, and/or publication of this article.
	
	\section*{Funding}
	The authors received no financial support for the research, authorship, and/or publication of this article.
	
	\section*{ORCID iDs}
	Ali Sabzejou \orcidlink{0009-0006-1095-3237}\\
	Farbod Azimmohseni \orcidlink{0009-0006-1858-6953}
	
	\section*{Supplementary Material}
	Supplemental material for this article is available online.
	
	\bibliographystyle{SageH}
	\bibliography{references.bib}
	
\end{document}